\documentclass[sigconf,nonacm]{acmart}

\AtBeginDocument{%
  \providecommand\BibTeX{{%
    \normalfont B\kern-0.5em{\scshape i\kern-0.25em b}\kern-0.8em\TeX}}}

\usepackage[para]{footmisc}
\usepackage{hyperref}
\usepackage{multirow}
\usepackage{multicol}
\usepackage{booktabs}
\usepackage{tablefootnote}
\usepackage{url}
\usepackage{graphicx}
\usepackage{subfigure}
\usepackage{listings}
\usepackage{color}
\usepackage{xcolor}
\definecolor{dkgreen}{rgb}{0,0.6,0}
\definecolor{gray}{rgb}{0.5,0.5,0.5}
\definecolor{mauve}{rgb}{0.58,0,0.82}
\usepackage{float}

\begin{document}

\title{GN-Transformer: Fusing Sequence and Graph Representation for Improved Code Summarization}

\author{Junyan Cheng}
\email{junyanch@usc.edu}
\affiliation{%
  \institution{University of 
Southern California}
  \city{Los Angele}
  \state{ca}
  \country{USA}
}

\author{Iordanis Fostiropoulos}
\email{fostirop@usc.edu}
\affiliation{%
  \institution{University of 
Southern California}
  \city{Los Angele}
  \state{CA}
  \country{USA}
}

\author{Barry Boehm}
\email{boehm@usc.edu}
\affiliation{%
  \institution{University of 
Southern California}
  \city{Los Angele}
  \state{CA}
  \country{USA}
}



%


\begin{abstract}

As opposed to natural languages, source code understanding is influenced by grammatical relationships between tokens regardless of their identifier name. Graph representations of source code such as Abstract Syntax Tree (AST) can capture relationships between tokens that are not obvious from the source code. We propose a novel method, GN-Transformer to learn end-to-end on a fused sequence and graph modality we call Syntax-Code-Graph (SCG). GN-Transformer expands on Graph Networks (GN) framework using a self-attention mechanism. SCG is the result of the early fusion between a source code snippet and the AST representation. We perform experiments on the structure of SCG, an ablation study on the model design, and the hyper-parameters to conclude that the performance advantage is from the fused representation. The proposed methods achieve state-of-the-art performance in two code summarization datasets and across three automatic code summarization metrics (BLEU, METEOR, ROUGE-L). We further evaluate the human perceived quality of our model and previous work with an expert-user study. Our model outperforms state-of-the-art in human perceived quality and accuracy.

\end{abstract}



\begin{teaserfigure}
  \includegraphics[width=\textwidth]{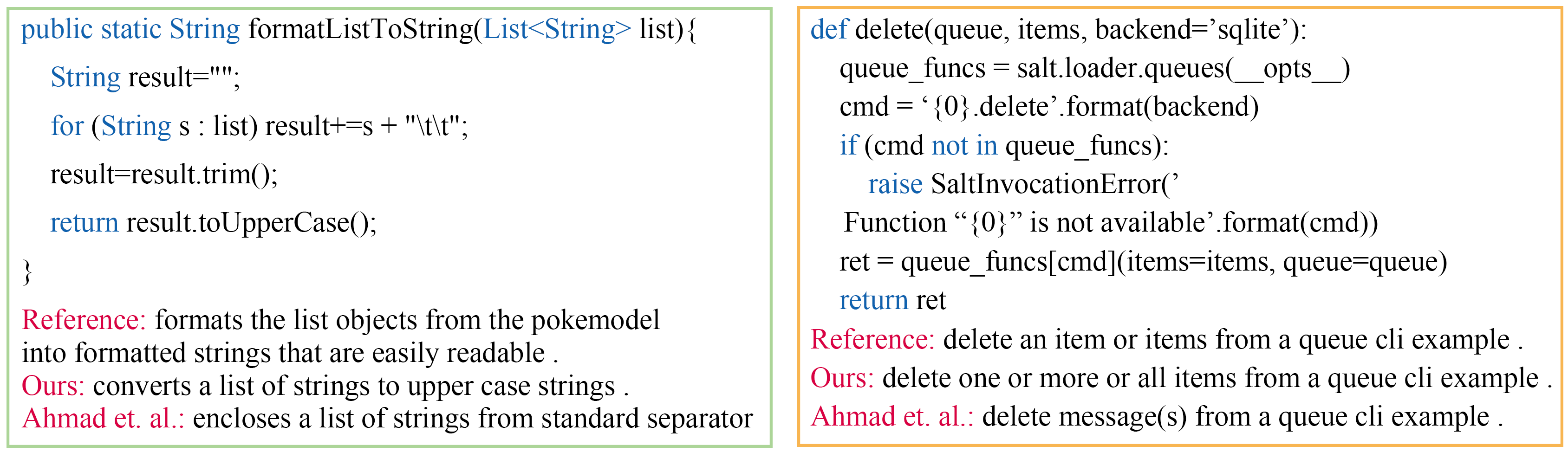}
  \caption{Examples of generated summaries on Java (green block) and Python (orange block) test sets. The comparison is between the dataset provided code-summary (Reference) and the code-summary generated for the same snippet by a GN-Transformer (Ours) and a Transformer by \citet{NCS}.}
  \label{fig:teaser}
\end{teaserfigure}

\maketitle


\section{Introduction}

\begin{figure*}[h]
\label{F_model}
\begin{center}
\includegraphics[width=0.75\textwidth]{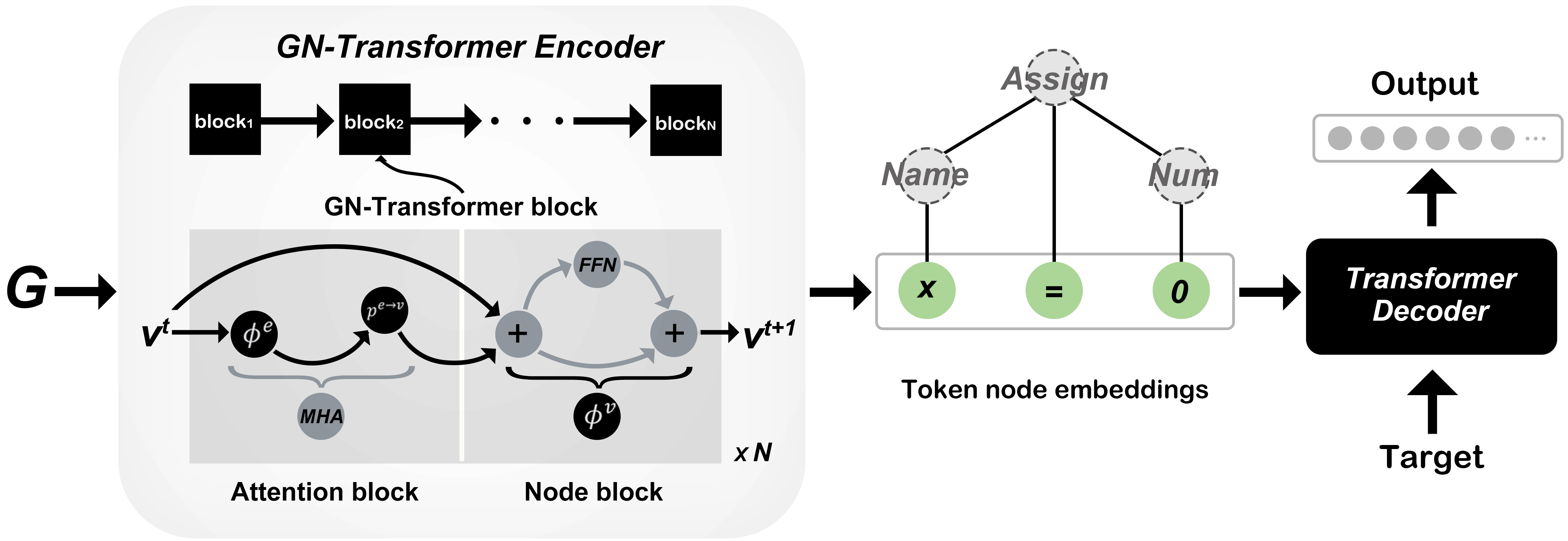} 
\end{center}
\caption{The encoder consists of multiple GN-Transformer blocks with a graph $G$ as the input. We denote `+' as a residual connection followed by a normalization layer. The Encoder outputs graph $G`$ with updated node attributes. For the task of code-summarization, only the token-nodes (green) are used as input to the decoder while AST-nodes (grey) are discarded.
} 
\label{figure:gnt_arch}
\end{figure*}

Code summarization is the task of generating a readable summary that describes the functionality of a snippet. Such a task requires a high-level comprehension of a source code snippet, and it is an effective way to evaluate whether a Deep Learning Model is able to capture information from complex code relationships. Programming languages are context-free formal languages. Abstract Syntax Tree (AST) is an unambiguous representation that can be derived from a source code snippet.
The structure of a snippet and the grammar relationships between the tokens can be described by an AST precisely and without noise. Such representation can provide an inductive bias for code understanding. 

Multiple methods perform code summarization and have used graph and tree representations using late fusion techniques. 
For example, \citet{CODE2SEQ} encoded AST paths using random walks, source code tokens using an LSTM and then fused them by an attention mechanism.
\citet{CGCNN} used Convolutional Neural Networks (CNN) to embed source code statements as node attributes in a Control Flow Graph (CFG). They apply DeepWalk \cite{DeepWalk} on the CFG to learn a representation which they concatenate with a source code representation embedded in a CNN-based model for the task of bug localization.
\citet{CodeGNN} apply Graph Convolutional Networks (GCNs) directly on an AST and a GRU on the source code sequence and concatenate their representations in a late fusion approach. 
The above methods propose different ways to extract features from an AST or CFG but with limited \textit{cross‐modal interaction} \cite{RS}. The graph and code modalities are fused late as the learned representations of independent models. This is opposed to training end-to-end on a modality that is an outcome of early fusion. Early fusion can improve model performance via cross-modal interactions \cite{RS}. 

Early fusion methods enable interactions between different modalities and allow a model to extract information of different granularity from each modality \citep{RS}.  
A graph is a flexible representation that allows for the early fusion of multiple modalities. Recent advancements in deep learning frameworks for graphs have helped the improvement of early fusion methods \citep{GN,MPNN,NLNN}. 

There are many approaches in learning representations of Graphs.
Spectral-based methods like GCN \citep{GCN} proposed graph convolution using the Laplacian of the adjacency matrix and is based on spectral graph theory. Spatial-based methods like GraphSAGE \citep{GraphSAGE} aggregate information from neighboring nodes using different aggregation functions like mean, summation, pooling, and LSTM.
However, there are several limitations of current approaches. Over smoothing problem \citep{oversmooth} makes it difficult to improve the model performance by increasing depth. Work by DGCN \citep{dgcn} proposes to solve the problem using skip connections. Identification of isomorphic graph structures is a difficult problem for graph models. The ability to identify isomorphism by a model corresponds to the ability to learn the graph structure. Work by GIN \citep{GIN} uses multi-layer perceptrons in a GNN framework to accurately identify isomorphism in graphs. 
When every node in a graph has equally weighted neighboring nodes it can add noise to the downstream task. GAT \citep{GAT} introduced an attention mechanism to learn a weight matrix for neighboring nodes. 

A Transformers architecture \citep{Transformer} combines all the above techniques using a multi-head attention mechanism and a feedforward network with residual connections. Thus a Transformers architecture can have a great advantage if formulated in the context of graphs. Recent work by \citet{GN} proposed a general graph deep learning framework, Graph Networks that can be applied in the context of multiple deep learning architectures. 
Our work is motivated by the recent advancements in graph networks and the advantages of Transformers architecture. 

BLEU and METEOR score are the dominant metric in evaluating deep learning model on the quality of machine generated text. ROUGE metric is used to evaluate the quality of text generated for summarization benchmarks. Previous work \cite{bleu_score,rouge_correlation} have shown that BLUE and ROUGE score do not correlate with human perceived quality. Metrics such as METEOR \cite{meteor} have been proposed as an alternative but do not evaluate the accuracy of the summary. Previous work in code summarization rely on automatic reported metrics and overlook weaknesses of such score in human perceived quality. 

In this paper we propose a novel architecture \textbf{GN-Transformer} shown in Figure \ref{figure:gnt_arch} and \textbf{Syntax Code Graph}, a fused AST and code sequence representation. Our model can learn end-to-end on an SCG under the Graph Network framework for the task of code summarization. We perform an ablation study on the model design as well as on two structure variants of a fused graph and sequence modality. We further evaluate our model using an expert-user survey on generated summaries. Based on our insights, we conclude that the performance advantage is from the GN-Transformer encoder that can efficiently learn the representation of an SCG. In detail:
\begin{itemize}

\item We extend Graph Networks (GN) 
to a novel \textit{GN-Transformer} architecture. A sequence of GN encoder blocks followed by a sequence-to-sequence Transformer decoder. 

\item We propose a novel method for the early fusion of a code snippet sequence and the corresponding AST representation called  \textit{Syntax-Code Graph} (SCG).

\item We evaluate our approach for the task of code summarization on two datasets. We use quantitative experiments and the largest to-date empirical study on code summarization. We compare our approach using three quantitative metrics (BLEU, METEOR, ROUGE-L), we corroborate the results with a large-scale survey of 330 participants and outperform previous state-of-the-art in both quantitative metrics and human perceived quality.
\end{itemize}

We evaluate our model on a Java \cite{TLCS} and Python \cite{PCSD} dataset. We compare our results to those of \citet{NCS} under identical experimental setups. Two qualitative results are presented in Figure \ref{fig:teaser}. Our code, trained models, and pre-processed dataset are available in \url{https://github.com/chengjunyan1/GN-Transformer-AST}.


\section{Related Work}
\label{others}

We identify two main categories of related works. Works that focus on the fusion between a sequence and graph modalities and works that focus on training end-to-end on a graph structure. 

Augmenting a source code modality with graph representations like AST and Control-Flow-Graph (CFG) has been used by deep learning techniques for source code understanding. 

Previous methods consider sequences and graphs as two modalities that are processed independently. For a sequence, recurrent architectures such as RNNs, LSTMs, GRUs are commonly used. 
Early fusion methods use domain knowledge to design a fused modality that enables cross-modal communication. As a result, late fusion approaches, are simple to implement but potentially lose their ability to exploit cross‐modal information \cite{RS}. 
Late fusion approaches consider the source code with the corresponding graph modality as input to two separate models. \citet{Code2Vec} extract AST embedding with random walks, concatenate it with the source code token embedding, and finally applies an attention mechanism, learning a context vector used for downstream tasks. \citet{DeepCom} propose Structure-Based Traversal (SBT) that flattens the AST into a sequence. \citet{CGCNN} apply DeepWalk with CNN and LSTM to learn a CFG representation which is then concatenated with the source code representation. \citet{CodeGNN} use GCN to learn the AST embedding which is then concatenated with source code embedding. 

\citet{FAAST} augment the AST edges using the `type' of the relationship between nodes extracted from the control and data flow graphs. A gated GNN is applied to train on the augmented AST representation. Such approaches are not able to effectively capture cross-modal interactions that can further improve the model performance \cite{MML}. There are no interactions between different modalities when encoding them with late fusion approaches \cite{RS}. 

There are some recent works that propose the early fusion of graph and source code representations for deep learning models.
\citet{NCS} propose the fusion of a graph into a sequence. They flatten an AST representation using SBT from \citet{DeepCom} and then use the sequence into a relative positional encoding and a copy mechanism on a Transformer architecture. Their approach resulted in performance degradation when compared to simply using the source code tokens as a sequence.
\citet{allamanis} propose a `program graph' which was based on AST. The graph is constructed by introducing edges designed by expert knowledge into the AST like ``LastWrite'', ``ComputedFrom''. The edges define relationships between tokens, the type of edge is applied using a rule-based approach. \citet{GREAT} propose to use relation-aware attention \cite{RPE} to incorporate graph information into attention computation. They use a ``leaves-only'' graph that is specific to the context of C language. For their graph structure, they discard all nodes that do not correspond to a source code token and introduce them as edges between tokens. The implementation of the graph structure is not publicly available. In addition, it relies on a complex language-specific preprocessing pipeline for C language, which makes it difficult to be extended to other programming languages. Our method relies only on AST which is easily available for all programming languages.

There are methods besides code understanding that incorporate graph or tree information into the sequence.
In natural languages, \citet{TreeTransformer} uses the constituency tree from sentences in the self-attention layer. TreeLSTM \citep{TreeLSTM} fuse tree information with a sequence using a tree-structured LSTM. Such approaches are limited only to tree structures. \citet{TextGCN} construct heterogeneous graphs of documents and text. A document contains multiple words and a word may appear in multiple documents. Many-to-many relationships are constructed between documents and words in a text. The constructed graph requires placing test documents with training document nodes, which is not suitable for the prediction of new unseen texts.


Recent work try to evaluate the correlation of automatic metric for code summarization with human perceived quality. \citet{stapleton2020human} compare code generated summaries with the reference descriptions in a human study of forty-five expert-user. They ask users to complete coding tasks using the provided summaries and evaluate their performance. The results show no correlation between human performance and automatic metrics. The study does not directly evaluate perceived quality and does not compare machine generated summaries from multiple models. 
\citet{gao2020generating} propose a sequence-to-sequence attention model specific for the task of generating a title for a given code snippet for a popular website, Stack Overflow. The study evaluated on automatic metric (BLUE, ROUGE) and the results are corroborated with a human study on perceived quality. However, the study was limited to five evaluators. 
\citet{panthaplackel2020learning} propose a novel task of modifying documentation for a given code snippet given the commit difference. They perform a human study in a repository commit simulation environment and ask users to discard or update source code documentation with a machine generated one. Modified annotations are generated from the proposed model and the baselines for a cross-model comparison. However, the number of participants was limited to nine university students. 

Our work is motivated by the advantage of early fusion, the lack of a flexible fused code representation and a model capable of learning on the fused modality end-to-end.

\section{Background}
\label{FS}

We discuss Graph Networks in Section \ref{sec:gn}. In Section \ref{cross_modal_interactions} we discuss the theoretical implications for improving cross-modal interactions using early fusion. In Section \ref{relational_inductive_bias} we discuss relational inductive biases that provide an advantage in representing a sequence in a graph instead of the contrary.

\begin{figure*}[h]
\label{F_fc_ast}
\centering
\subfigure[]{
\begin{minipage}[t]{1\columnwidth}
\includegraphics[width=0.75\linewidth]{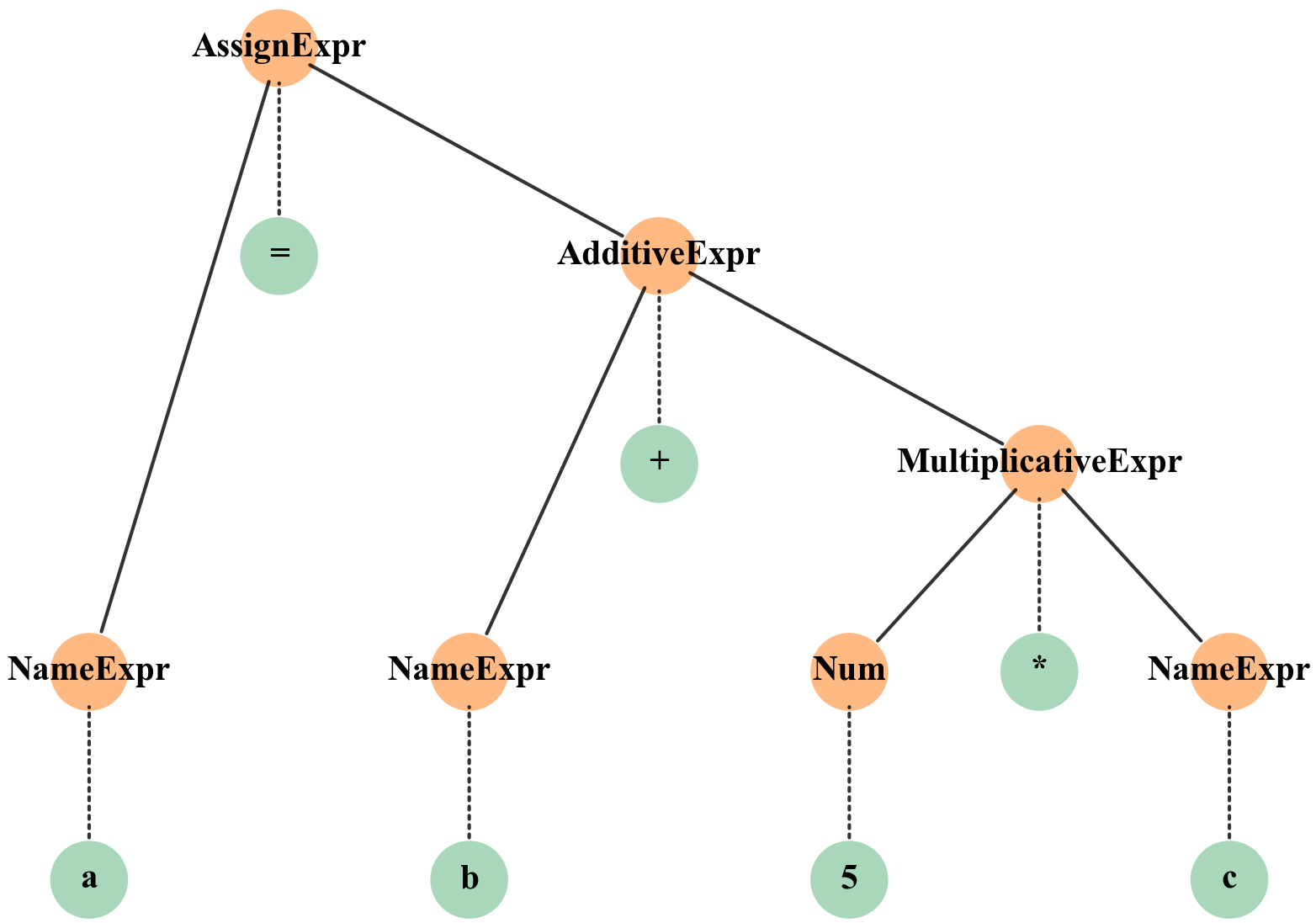}
\end{minipage}
}
\subfigure[]{
\begin{minipage}[t]{1\columnwidth}
\includegraphics[width=0.75\linewidth]{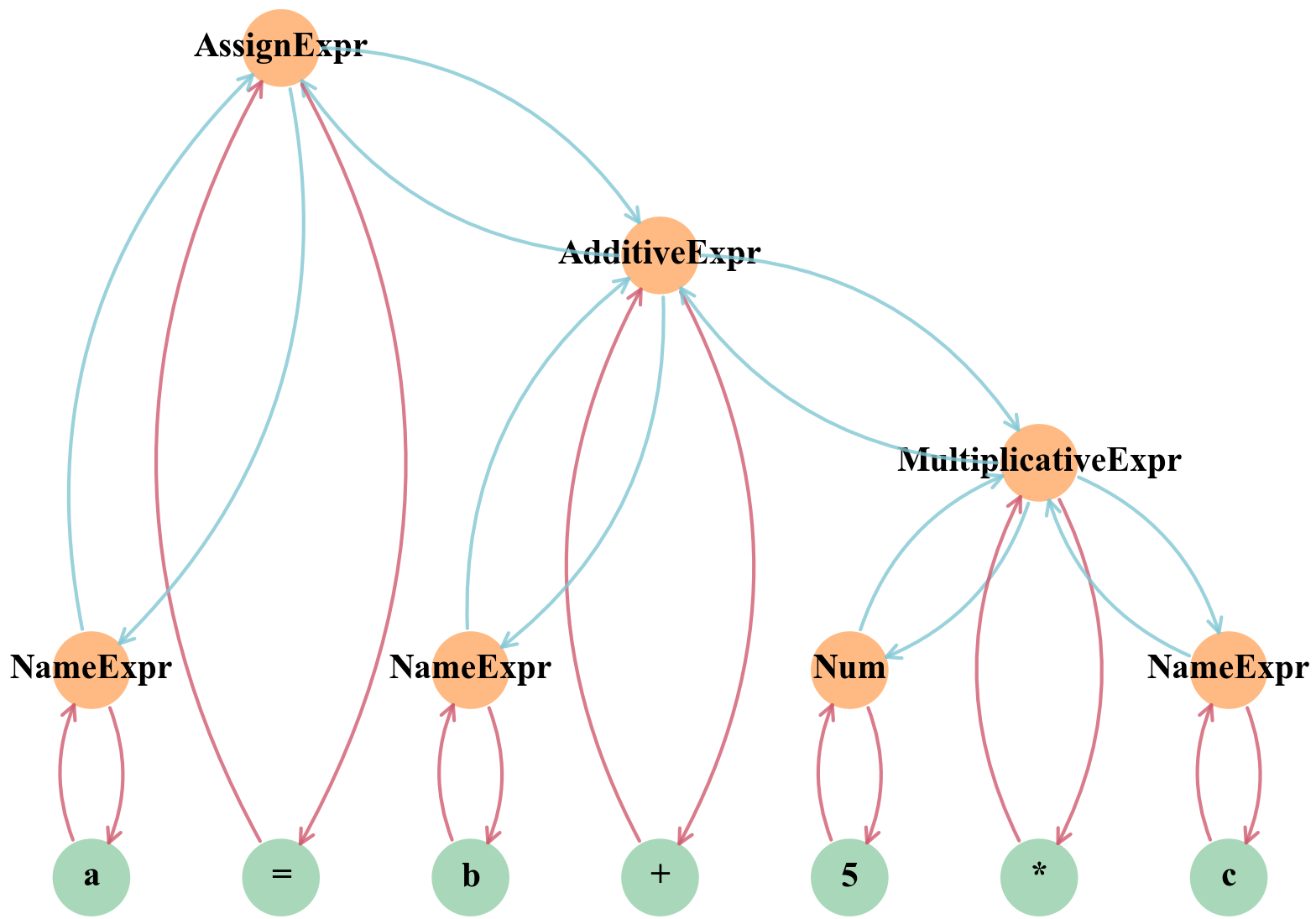}
\end{minipage}
}
\centering
\caption{
(a) Orange nodes denote an AST for the statement `a=b+5*c', dashed lines connect each token-node (green) with their direct parent in AST. (b) Standard SCG structure preserve the AST structure with additional edges between AST-nodes and tokens. }
\label{figure:ast_scg}
\end{figure*}

\subsection{Graph Networks} 
\label{sec:gn}

Graph Networks \citep{GN} is a general framework that unifies deep learning models with a graph representation. Graph Networks is a generalization of previous approaches including Message Passing Neural Networks \cite{MPNN}, Non-Local Neural Networks \cite{NLNN} that can learn a graphical representation of data under a configurable computation unit, GN block. 

A GN block can flexibly define computations on a graph by configuring subblocks that are a composite of functions. A GN block is composed of three configurable subblocks. A \textbf{Node block} updates the node attributes, an \textbf{Edge block} updates the edge attributes then aggregates edge information, and a \textbf{Global block} maintains a global graph attribute by aggregating all nodes and edges information on the graph. 

By adding, removing, or combining different subblocks, we can flexibly define arbitrary computation rules on a graph. 
In our work, we use a Node block and a modified Edge block, which we call Attention block, to define a single GN-Transformer block. We discuss in further detail our method in Section \ref{GA_block}.

The GN framework provides flexibility in the structure of the graph representation. The definition of nodes and edges and their relationships are context dependant. For example, a node could be a word in a sentence \citep{Treebank}, local feature vector in a CNN’s output feature map \citep{dgcnn}, or balls and walls for modeling a rigid body system \citep{rigid}. An edge could represent the energy in a physical system or the relative position for image patches. 

At the end of each GN layer, the interaction between nodes is through the information propagation by the \textit{Node block}. One GN layer is the execution of one round of information propagation on the graph. The blocks within each layer define the rules of how information propagation on the graph will happen. In our method, we configure the node block and modify the edge block as an attention mechanism block, corresponding to 1-hop information propagation. Thus, an N-block GN-Transformer Encoder corresponds to N-hop information propagation between graph nodes.

Using Graph Networks frameworks, we can represent data with arbitrary relational inductive biases which will discuss in Section \ref{relational_inductive_bias}. Moreover, a graph modality can provide flexibility as a fusion unification framework to represent multiple sources of information. 

\subsection{Cross-Modal Interactions} 
\label{cross_modal_interactions}

Multi-modal signals exhibit exploitable correlations which are defined as \textit{cross-modal interactions}. Cross-model information enables the model to augment the feature extraction for individual modality through other modalities implicitly \citep{RS}.

Early fusion achieves improved interactions between modalities. Graph Networks can be regarded as a general framework for early fusion. Due to the representation power of the graph, it is possible to represent different modalities within a graph. Representing multiple modalities on a single graph, then using the information propagation attribute of GN the interaction between modalities is implicitly learned. For example, to fuse the information of an image and the caption, we can define one set of nodes representing extracted image features or even image patches, and another set of nodes representing token embedding of the caption. Consequently, the information in the image could directly interact with the information contained in the text. The cross-modal interaction could happen from a low-granularity raw data input such as pixels to high-granularity abstract features such as features extracted by convolutional filters.

\subsection{Relational Inductive Bias} 
\label{relational_inductive_bias}

Relational inductive biases impose constraints on the relationship and interactions among entities in a learning process. In Graph Networks, the assumption that two entities have interaction is expressed by an edge between the corresponding entity nodes. The absence of an edge expresses the assumption of isolation, which means there is no direct interaction between the two nodes.

Sequences are unstructured data in which relationships are implicit. Implicit relationships can be represented with each token as a graph node that is fully connected with all other nodes. Such representation allows each node to interact with every other node with no explicit assumptions on isolation. Thus it allows the model to infer the relationships between them implicitly. Transformers could be regarded as inferring on such a fully connected attention graph. Each token in an input sequence corresponds to a node. The relationship between tokens is thus represented by attention weights, high attention values correspond to strong interactions, while low attention means isolation.
 
However, as shown by the recent work \citet{ViT}, a model without inductive biases may perform worse than the model with valuable inductive biases when the dataset is relatively small. It is less efficient for a model to learn to infer relationships without any explicit assumptions of interactions and isolation \citep{GN}.
AST provides precise information about interactions and isolation among tokens in source code since it brings information about the grammatical relationships between tokens. 
We can find an explicit mapping between tokens in a sequence and nodes in the AST through the scope information provided by a parser for a given programming language. 
We further discuss the graph structure of input data and how we fuse AST with it in Section \ref{GS}. We can introduce the relational inductive bias of AST into our model through the fused graph representation.

There are several benefits in fusing the sequence with the graph representation as opposed to flattening a graph representation into a sequence. Firstly, a graph representation can contain information sources with arbitrary explicit or implicit relational inductive bias through the graph structure. Secondly, a graph structure can be augmented using expert knowledge \cite{GN}, the `program graph' from \citet{allamanis} could be regarded as an example that introduces expert knowledge. They use hand-crafted edges into an AST to improve performance in the variable misuse task. Thirdly, a graph representation results in better combinatorial generalization \citep{GN} due to the reuse of multiple information sources simultaneously in a unified representation. As an example consider two code snippets that have the same functionality but different variable names. The AST for these snippets will be the same, as such the model could use the name invariant AST representation to generalize better on seemingly different code snippets.


\section{Method}
\label{GS}

In Section \ref{ast_code} we propose a joint graph representation of source code and AST called standard Syntax-Code Graph (SCG). In Section \ref{Enc-Dec} we introduce our model architecture and propose our Graph Networks based encoder in Section \ref{GA_block}. Our model follows the generic Transformer encoder-decoder architecture. The encoder is an extended architecture of Graph Networks with multiple GN blocks. The decoder is a vanilla Transformer decoder. The overview of our architecture is presented in Figure \ref{figure:gnt_arch}.

\subsection{Syntax-Code Graph}
\label{ast_code}

\textbf{Syntax-Code Graph} (SCG) is the fused graph representation between an AST and a source code sequence. 
The graph is built based on the AST, with source code tokens as additional nodes, and additional edges connecting them with the AST. We call the nodes on the AST as \textit{AST-node}, the source code tokens as \textit{token-node}.
The node attribute of an AST-node is the identifier name on the AST, such as ``NameExpr'', ``AssignExpr''. The node attribute of a token-node is the source code text associated with that token, such as ``a'', ``int'', ``+''. 

We connect token-nodes with AST-nodes using their \textit{direct parent} in an AST. Each AST-node has a scope provided by an AST parser that corresponds to a range of positional marks in the original source code snippet. Each positional mark is a line and column pair.  We define the direct parent as the deepest AST-node that includes the token-node in their scope. Consider the statement `a=b+5*c' as an example, the AST is shown in Figure \ref{figure:ast_scg}(a). `MultiplicativeExpr' has a positional mark \textit{line 1, col 5 $\sim$ line 1, col 7} which corresponds to `5*c', `Num' and `NameExpr` have positional marks of \textit{line 1, col 5 $\sim$ line 1, col 5} and \textit{line 1, col 7 $\sim$ line 1, col 7} respectively. Token-nodes `5' and `c' in this case belong to the scope of multiple AST-nodes (e.g., `AdditiveExpr', `AssignExpr'), but their direct parents are `Num' and `NameExpr' respectively. The dashed lines in Figure \ref{figure:ast_scg}(a) connect each token-node with their direct parent AST-node. The graph on Figure \ref{figure:ast_scg}(b) shows the final SCG.

We perform two ablation studies on the structure of SCG to see the effectiveness of connecting token-nodes with only their direct parents and the effects of isolation between token-nodes by fully connecting them. Results can be found in Section \ref{sec:abl_scg}.



\subsection{Encoder-Decoder Architecture}
\label{Enc-Dec}

The encoder consists of a stack of $N$ GN-Transformer blocks which are derived from GN blocks, the main computation unit in Graph Networks, it takes a graph as input and returns a graph as output \citep{GN}. Each block in our model implements a Multi-Head Attention layer and a Feed-Forward Neural sublayer equivalent to the Transformer encoder sub-layers. Encoder accepts a graph $G=\{V,E\}$ where $V$ is a node set containing the set of initial node attributes $h^0 \in \mathbb{R}^{\left | V \right |\times d_{model}}$ where $d_{model}$ is the input and output dimension of the encoder. $E$ is the edge set used to identify the neighboring node set $N_i$ for each node $i$. The node features in the input graph are initialized through an embedding layer. AST and token nodes fetch an embedding vector according to their types and identifier names respectively. For our implementation, we handle token-node identifier and AST-node type separately when performing an embedding lookup. This is to avoid representing the same embedding nodes of different types with a naming conflict.
The encoder outputs the graph with the updated node features $h^N$. Feature vectors of token-nodes are fetched as the decoder input. Feature vectors of AST-nodes are discarded and the token embeddings are padded for batching (see Figure \ref{figure:gnt_arch}). 

\subsection{GN-Transformer Blocks}
\label{GA_block}

\textbf{GN-Transformer block} is an extension to GN blocks as proposed by \citet{GN} with Multi-Head Attention (MHA) and Feed-Forward Network (FFN) defined in the context of Graph Networks. As discussed in Section \ref{sec:gn}, in the context of GN, a GN-Transformer block will execute one round of information propagation and aggregation on the neighboring nodes that update the node attributes in the graph.
The $t$-th GN-Transformer block accepts node attributes $h^t$ as input and output the updated node attributes $h^{t+1}$. 
The GN-Transformer block is composed of two subblocks, an \textit{Attention block} and \textit{Node block}. The information propagation is an outcome of the two subblocks applied sequentially. 

An \textbf{Attention block} implements Multi-Head Attention \textbf{MHA}. It is composed of an \textbf{attention update function} $\phi^e$ which calculates the attention weight between each node and their neighboring nodes, and an \textbf{attention aggregate function} $p^{e\rightarrow v}$ which aggregates the information for each node from all of their neighboring nodes weighted by the attention weights, denoted as attention $head^{(\gamma)}_i$. 

The unnormalized attention weight between nodes $i$ and $j$ from $\gamma$-th attention head for the $t$-th GN-Transformer block is $\alpha^{(\gamma)}_{ij}$. Each Attention block has $H$ attention heads. Where $W_{\gamma}^{K}, W_{\gamma}^{Q} \in \mathbb{R}^{d_{model} \times d_{k}}$ are parameter matrices for attention head $\gamma$. The attention weight from node $i$ to node $j$ for attention head $\gamma$ is computed by the attention update function $\phi^e$ as follows:\[
\alpha^{(\gamma)}_{ij}=\frac{h^t_{i}W_{\gamma}^{Q}(h^t_{j}W_{\gamma}^{K})^\textit{{T}}} {\sqrt{d_{k}}}
\]
Subsequently, the attention aggregate function $p^{e\rightarrow v}$ aggregates the information between each node $i$ and the set of neighboring nodes $N_i$ based on the attention weight computed above as:
\[ \bar{a}^{t+1}_i=Concat(head_{i}^{(1)},...,head_{i}^{(H)})W^{O}\]
Where \[
head_{i}^{(\gamma)}=\sum_{j \in N_{i}}h^t_{j}W_{\gamma}^{V}\sigma(\alpha _i^{(\gamma)})_j
\]
and $\sigma(\cdot)$ is the softmax function, with $W_{\gamma}^{V} \in \mathbb{R}^{d_{model} \times d_{v}}$, $W^{O}$ as parameter matrices. 

A \textbf{Node block} is composed of a \textbf{node update function} $\phi^v$, that updates the attribute of each node with the aggregated information of all of their neighboring nodes $\bar{a}^{t+1}_i$.
The output of a GN block is the updated node attributes $h^{t+1}$ updated by $\phi^v$ using the equation:
\begin{center}
$h^{t+1}_i=FFN(h^t_i+\bar{a}^{t+1}_i)+h^t_i+\bar{a}^{t+1}_i$
\end{center}
A Feed-Forward Network \textbf{FFN} with a residual connection is used such as a single layer MLP with a non-linearity such as \textit{ReLu}. For our experiments, we also use dropout and layer normalization identical to \citet{Transformer}. An illustration of our model can be found in Figure \ref{figure:gnt_arch}.


\begin{figure*}[h]
\centering
\subfigure[]{
\begin{minipage}[t]{.51\columnwidth}
\includegraphics[width=1\linewidth]{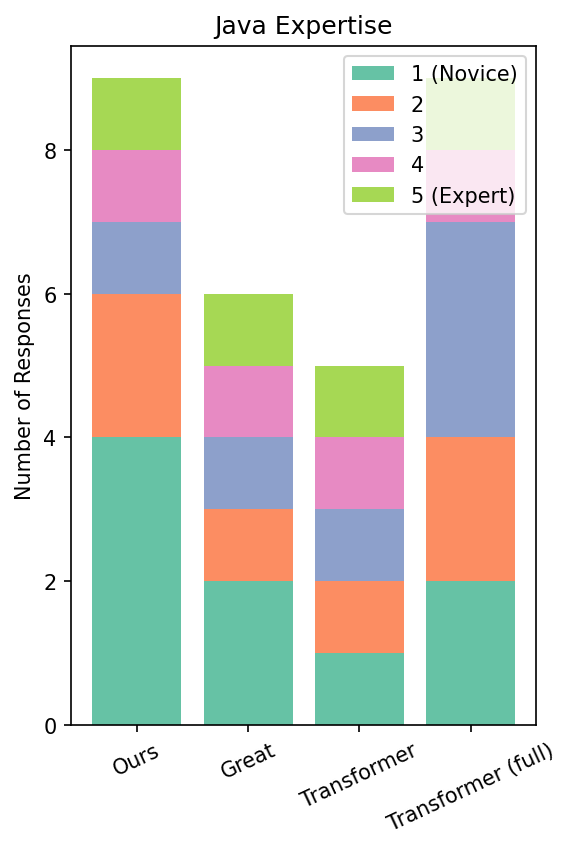}
\end{minipage}
}
\subfigure[]{
\begin{minipage}[t]{.48\columnwidth}
\includegraphics[width=1\linewidth]{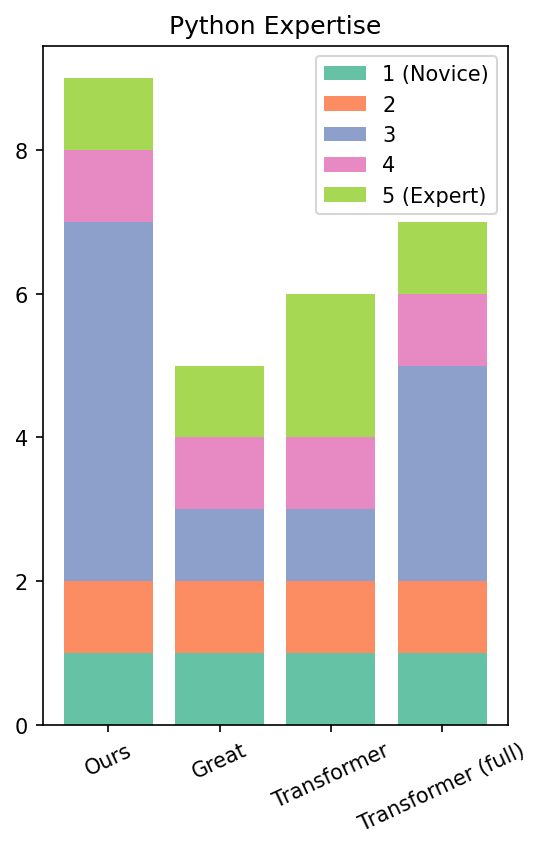}
\end{minipage}
}
\subfigure[]{
\begin{minipage}[t]{.48\columnwidth}
\includegraphics[width=1\linewidth]{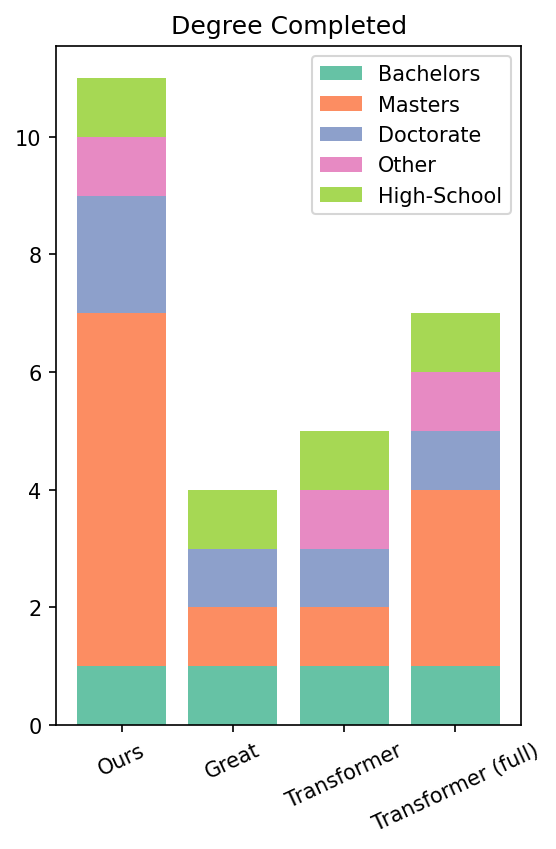}
\end{minipage}
}
\subfigure[]{
\begin{minipage}[t]{.48\columnwidth}
\includegraphics[width=1\linewidth]{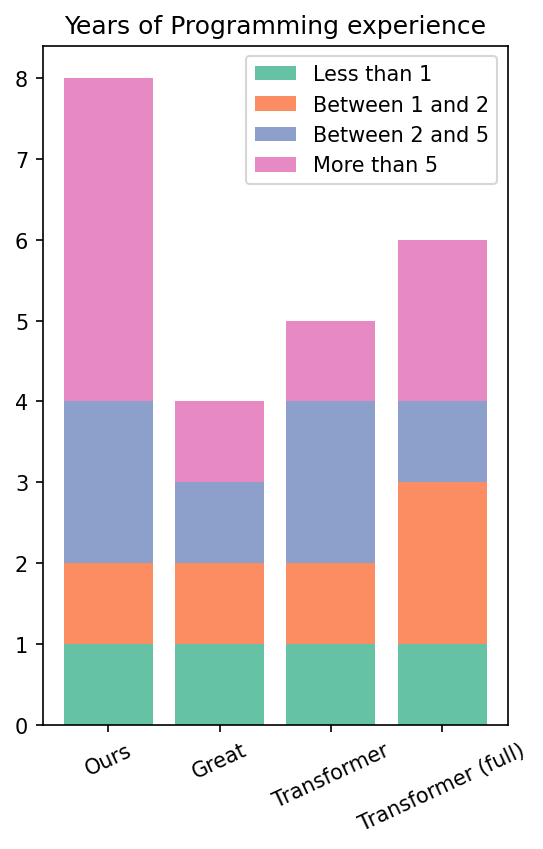}
\end{minipage}
}
\centering
\caption{ Survey participants were asked to choose the best description for a given code snippet. Comparison between GN-Transformer (``Ours''), \citet{GREAT} (``GREAT''), \citet{NCS} (``Transformer (full)'') and \citet{Transformer} (``Transformer''). Survey results were analyzed by the demographic distribution of participants on (a) self-reported Java expertise (b) self-reported Python expertise (c) highest level of education completed or currently in progress (d) years of programming experience. }

\label{fig:survey}
\end{figure*}
\section{Experiment}
\label{EX}

We evaluate our model on two code summarization datasets. We perform experiments on the hyperparameters, model structure, variants of the graph structure and perform a user study. Our experiment settings are presented in Section \ref{ex_set}. Results are presented and analyzed in Section \ref{ex:cs}, \ref{sec:abl_java} and \ref{sec:abl_scg}.

\subsection{Settings}
\label{ex_set}

The experiments are conducted on a Java dataset \citep{TLCS} and a Python dataset \citep{PCSD}.
Our preprocessed datasets are composed of the source code, corresponding AST, and a text summary. 
We used JavaParser\footnote{\url{https://javaparser.org/}} to extract the AST and javalang\footnote{\url{https://github.com/c2nes/javalang}} for parsing Java source code, python ast\footnote{\url{https://docs.python.org/3/library/ast.html}} to parse and get the AST for Python. 

We chose a Transformer as our main comparison baseline which achieved state of the art in the two datasets. \citet{NCS} propose a base model which is a vanilla sequence-to-sequence Transformer and an extended model (full) that uses relative positional embedding and a copy mechanism. We perform experiments where we reproduce their results on our pre-processed datasets, using only the source code tokens as input to their model. 
We additionally compare our method with \citet{GREAT}. Since their graph structure is not publicly available, we reproduced their method on SCG for comparison. 
We also compare our method with the results of other baselines reported in \citet{NCS}. We use metrics BLUE, ROUGE-L, and METEO that are also reported by all baselines. We apply the same hyper-parameters as \citet{NCS} which are listed in Table \ref{table:hp}.

\begin{table}[!htb]
\begin{center}
\caption{Summary of hyper parameters.}
\begin{tabular}{lll}
\multicolumn{1}{c}{\bf Hyper-parameters}  &\multicolumn{1}{c}{\bf Value} 
\\ \hline \\
Num of layers           &6\\
Attention heads         &8\\
$d_{k}$                 &64\\
$d_{v}$                 &64\\
$d_{model}$             &512\\
Hidden units in FFN     &2048\\
Dropout rate            &0.2\\
Optimizer               &Adam\\
Initial learning rate   &0.0001\\
Decay rate              &0.99\\
Max epoch num           &200\\
Early stop epochs       &20\\
Training Batch set      &30\\
Testing Beam size       &4\\
Max src. vocab size     &50000\\
Max tgt. vocab size     &30000\\
Max Java training code len.        &150\\
Max Python training code len.      &400\\
Max Java training summary len.     &50\\
Max Python training summary len.   &50\\ \hline \\
\end{tabular}
\label{table:hp}
\end{center}
\end{table}

For data preprocessing, the source code and summary data are truncated if they exceed the maximum length. We also discard all AST-nodes if their scope contains a discarded token-node. We set a vocabulary size limit, and we store only the highest frequent words. Words that are not in the vocabulary will be replaced by a special token $Unknown\_word$. Our methodology is consistent with \citet{NCS}.

\begin{table}[!htb]
\caption{Statistics of preprocessed datasets.}
\begin{center}
\begin{tabular}{lll}
\multicolumn{1}{c}{\bf Statistics}  &\multicolumn{1}{c}{\bf Java}  &\multicolumn{1}{c}{\bf Python}
\\ \hline \\
Examples - Train        &69593  &64939\\
Examples - Validation   &8694   &21605\\
Examples - Test         &8689   &21670\\
Unique Function Tokens  &66569  &104839\\
Unique Summary Tokens   &46859  &64898\\
Avg. Function Length    &120.29 &132.64\\
Avg. Summary Length     &17.73  &9.56\\
Avg. AST Nodes          &50.30  &70.10\\
Avg. AST Edges          &45.89  &68.16\\ \hline \\
\end{tabular}
\end{center}
\label{table:data}
\end{table}

\begin{figure}[h]
\centering
\subfigure[]{
\begin{minipage}[t]{0.45\linewidth}
\includegraphics[width=1\linewidth]{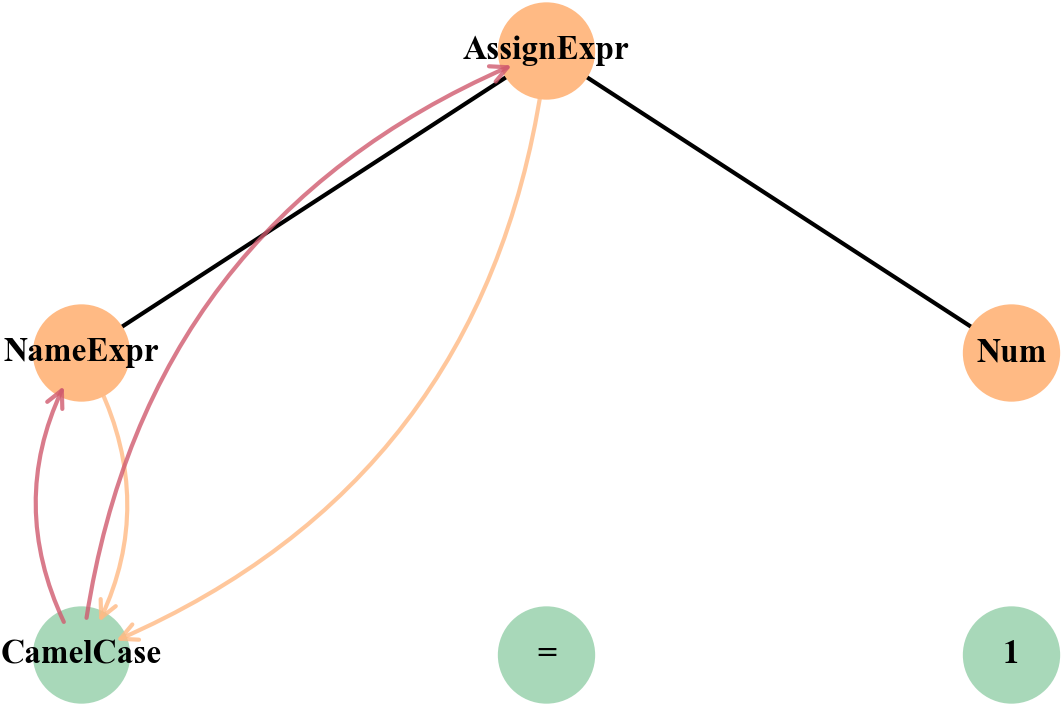}
\end{minipage}
}
\subfigure[]{
\begin{minipage}[t]{0.45\linewidth}
\includegraphics[width=1\linewidth]{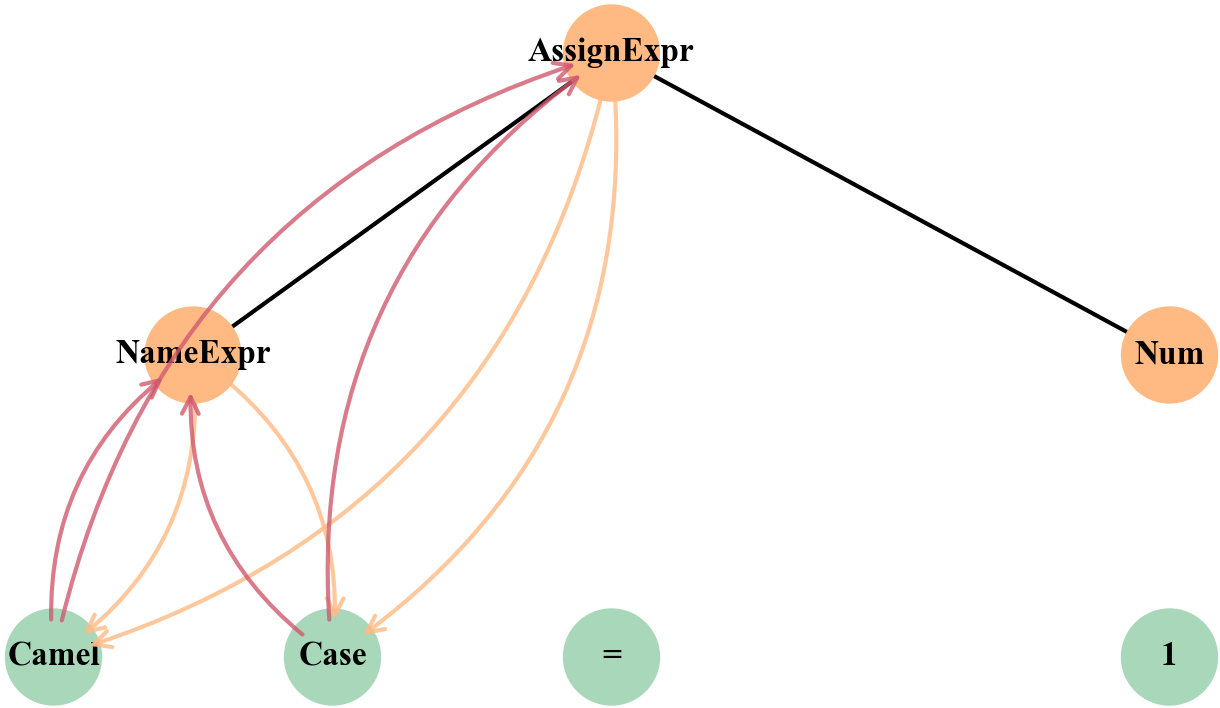}
\end{minipage}
}
\centering
\caption{Example of constructing edge connections for subtokens. Edges between other tokens and AST are omitted. Figure shows methodology for connecting shortcut edges for SCG Variant 1 for the subtoken-nodes. (a) Connection for original token-node. (b) Connection for subtoken-nodes. All subtoken-nodes copy the same edge from the original token-node.}
\label{figure:subtok_scg}
\end{figure}

The statistics of our pre-processed datasets are shown in Table \ref{table:data}. Despite the difference in implementation, we kept our methodology consistent with \citet{NCS}. We used the same $CamelCase$ and $snake\_case$ tokenizer from \citet{NCS} to preprocess source code data in both Java and Python datasets. SCG for subtokens can be found in Figure \ref{figure:subtok_scg}.  We also replace the strings and numbers in the source code with `$\langle STR\rangle$' and `$\langle NUM\rangle$'. For the code summary, we use the raw corpus for the Java dataset and use the same method with \citet{RL+Hybrid2Seq} to process code summaries in the Python dataset. We use the train/valid/test split from the original corpus for the Java dataset and we split the Python dataset by 6:2:2. All of the above are consistent with \citet{NCS}.

There are two differences between our preprocessed dataset and \citet{NCS}:

1. Data cleaning - We discard the samples that cannot be parsed by the compiler. There are 160 out of 87136 and 4894 out of 113108 samples in Java and Python datasets that are discarded, respectively. In contrast, there are no samples discarded by \citet{NCS} for the Java dataset. For the Python dataset, \citet{NCS} follow the same cleaning process by \citet{Wei} and discard 20563 out of the 113108 samples that exceed the length threshold. 

2. Python Dataset - We used the same methodology for preprocessing the Python dataset as we did for Java. In contrast, \citet{NCS} delete special characters in the Python source code while we preserve them. As a consequence, the average code length of their preprocessed dataset is 47.98 while ours is 132.64.

Apart from the above two differences, the preprocessing methodology is consistent with \citet{NCS} as well as other baselines reported by them.

\subsection{Code Summarization}
\label{ex:cs}

Table \ref{table:overrall_results} shows the comparisons of our model GN-Transformer with previous works for code summarization. Our method outperforms all previous works in all metrics.
The most suitable comparison of our approach with all previous works is that of a vanilla Transformer rather than a Transformer (full). Our work is best compared with a vanilla Transformer by \citet{NCS} as the main difference between the two architectures is fusing source code with a Graph as opposed to using only the code sequence, which is the scope of this paper. Transformer (full) makes use of additional improvements relevant only to sequence-to-sequence models such as positional embedding and copy mechanism introduced by \citet{COPY,RPE}. Such improvements do not directly apply to a GN-Transformer Encoder. We discuss potential interpretations of positional information on a graph in Section \ref{sec:pos_encoding} but leave any further analysis in the context of GN as future work. 

The result shows an improvement of 1.49, 2.09 BLEU, 0.51, 1.10 METEOR, and 1.99, 2.34 ROUGE-L in Java and Python datasets respectively. Although the improvement appear to be marginal and incremental it is consistent with improvement introduced by previous work. We find that the small improvement in the automatic metrics reported result in significant improvement for human evaluation metrics as concluded from the experiments in Section \ref{sec:exp_survey}.

\begin{table}[]
\caption{Overall results on Java and Python datasets.}
\begin{tabular}{cccc}
\hline
\textbf{Models}                & \textbf{BLEU}           & \textbf{METEOR}         & \textbf{ROUGE-L}        \\ \hline
\multicolumn{4}{c}{\textbf{Java}}                                                 \\ \hline
CODE-NN \cite{CODE-NN}             & 27.60          & 12.61          & 41.10          \\
Tree2Seq \cite{TreeLSTM}           & 37.88          & 22.55          & 51.50          \\
RL+Hybrid2Seq \cite{RL+Hybrid2Seq} & 38.22          & 22.75          & 51.91          \\
API+CODE \cite{TLCS}               & 41.31          & 23.73          & 52.25          \\
DeepCom \citep{DeepCom}                & 39.75          & 23.06          & 52.67          \\
Dual Model \cite{DualModel}            & 42.39          & 25.77          & 53.61          \\
GREAT \citep{GREAT}             &44.05          &26.42              &53.01  \\
Transformer \cite{NCS}            & 43.99          & 26.40          & 53.30          \\
Transformer (full) \cite{NCS}     & 44.84          & 26.89          & 54.80          \\
\textbf{Ours} & \textbf{45.48} & \textbf{26.91} & \textbf{55.29} \\ \hline
\multicolumn{4}{c}{\textbf{Python}}                                               \\ \hline
CODE-NN \cite{CODE-NN}             & 17.36          & 9.29           & 37.81          \\
Tree2Seq \cite{TreeLSTM}           & 20.07          & 8.96           & 35.64          \\
RL+Hybrid2Seq \cite{RL+Hybrid2Seq} & 19.28          & 9.75           & 39.34          \\
API+CODE \cite{TLCS}               & 15.36          & 8.57           & 33.65          \\
DeepCom \citep{DeepCom}               & 20.78          & 9.98           & 37.35          \\
Dual Model \cite{DualModel}           & 21.80          & 11.14          & 39.45          \\
GREAT \citep{GREAT}             &31.19          &18.65              &43.75  \\
Transformer \cite{NCS}          & 31.22          & 18.56          & 44.22          \\
Transformer (full) \cite{NCS}     & 32.79          & 19.63          & 46.51          \\
\textbf{Ours} & \textbf{33.31} & \textbf{19.66} & \textbf{46.56} \\ \hline
\end{tabular}
\label{table:overrall_results}
\end{table}

\begin{table*}[!htb]
\caption{Additional experimental results. Unlisted values are identical to the base model. Parameter number not including the embedding layer.}
\begin{center}
\begin{tabular}{c|cccccc|ccc|c}
\hline
\multicolumn{1}{l|}{} & $N$ & $d_{model}$ & $d_{ff}$ & $H$ & $d_k$ & $d_v$ & BLEU  & METEOR & ROUGE-L & \begin{tabular}[c]{@{}c@{}}params\\ ($\times10^6$)\end{tabular} \\ \hline
base                  & 6   & 512         & 2048     & 8   & 64    & 64    & 45.48 & 26.91  & 55.29   & 44.1                                                            \\ \hline
\multirow{11}{*}{(A)} & 2   &             &          &     &       &       & 37.37 & 20.66  & 49.20   & 14.7                                                            \\
                      & 4   &             &          &     &       &       & 43.41 & 24.86  & 53.62   & 29.4                                                            \\
                      & 8   &             &          &     &       &       & 45.76 & 27.18  & 55.59   & 58.9                                                            \\ \cline{2-11} 
                      &     &             &          & 1   & 512   & 512   & 43.13 & 24.77  & 53.24   &                                                                 \\
                      &     &             &          & 4   & 128   & 128   & 44.79 & 26.20  & 54.59   &                                                                 \\
                      &     &             &          & 16  & 32    & 32    & 45.52 & 27.00  & 55.39   &                                                                 \\
                      &     &             &          & 32  & 16    & 16    & 45.50 & 27.11  & 55.49   &                                                                 \\ \cline{2-11} 
                      &     & 256         &          &     &       &       & 40.01 & 22.35  & 51.16   & 22.1                                                            \\
                      &     & 1024        &          &     &       &       & 46.19 & 28.02  & 55.86   & 88.2                                                            \\ \cline{2-11} 
                      &     &             & 1024     &     &       &       & 43.99 & 25.40  & 54.34   & 31.5                                                            \\
                      &     &             & 4096     &     &       &       & 45.82 & 27.63  & 55.71   & 69.3                                                            \\ \hline
\multirow{3}{*}{(B)}  & \multicolumn{6}{c|}{2-hop GN-Transformer block}                 & 43.90 & 25.24  & 53.49   & 50.4                                                            \\
\multicolumn{1}{l|}{} & \multicolumn{6}{c|}{GAT}                           & 38.51 & 21.14  & 48.39   & 26.8                                                            \\
\multicolumn{1}{l|}{} & \multicolumn{6}{c|}{GAT with FFN}                  & 40.60 & 22.71  & 50.05   & 39.4                                                            \\ \hline
\multirow{2}{*}{(C)}  & \multicolumn{6}{c|}{Variant 1}                     & 45.25 & 26.56  & 55.11   &                                                                 \\
                      & \multicolumn{6}{c|}{Variant 2}                     & 44.42 & 26.87  & 54.08   &                                                                 \\ \hline
\end{tabular}
\end{center}
\label{table:add_exp}
\end{table*}

\subsection{Ablation Study on SCG}
\label{sec:abl_scg}

We perform an ablation study on two variants of SCG to study the effectiveness of the SCG structure, results are discussed in Section \ref{sec:abl_java}.
In \textbf{Variant 1} of SCG, we introduce shortcut edges between each AST-node and the token-nodes that are related to this AST-node. The distance between AST-nodes and all nodes within their scope is shortened to 1. It does not break any isolation among token-nodes but introduces additional direct interactions between token-nodes and AST-nodes. 
We define the related token-nodes of an AST-node as all token-nodes within the scope of the AST-node. 

In \textbf{Variant 2} of SCG, we make the token-nodes fully connected, which can be regarded as identical to the information passing in a Transformer. The nodes are not isolated from each other anymore and can interact with each other. The difference with a vanilla Transformer is that we have supplementary AST information. 


\subsection{Ablation study results on Java dataset}
\label{sec:abl_java}

We conduct an ablation study on three aspects of the model design using the Java dataset. Results are shown in Table \ref{table:add_exp}.

(A) Hyperparameters. 
Experiments on different combinations of hyperparameters including the number of layers, embedding size, width of FFN, and configurations of attention heads. 

(B) Model structure. 
We test the effectiveness of the Node block. We use a 2-hop GN-Transformer block with two MHA sublayers in each block thus aggregating two hops of information. The results show that the lack of Node aggregate function harms the performance. Collecting two hops of information without an FFN is less expressive. 
To determine the effectiveness of the GN-Transformer block, we replace it with a Graph Attention Networks (GAT) layer using the same hyperparameters as that of our base model. 
GAT did not perform optimally in our problem when modeled text sequence as a graph. The results show that the GN-Transformer block largely outperforms GAT for configurations with similar parameters as in (A). We add an FFN to GAT layers for an ablation study on MHA. Results show that MHA in combination with FFN brings a significant improvement. We thus conclude that both the FFN and MHA used by our GN-Transformer blocks are necessary and greatly improve performance. 

(C) Variants of graph structures.
We tested two variants of graph structures discussed in Section \ref{sec:abl_scg}. Both variants underperform when compared to an SCG. The performance of Variant 1 is much closer to standard SCG compared to Variant 2. Variant 1 introduces redundant edges and leads to redundant interactions between the AST and tokens. It slightly degrades performance since the structural information and isolation among tokens introduced by the AST are preserved. The results demonstrate the effectiveness of our design of only connecting token-nodes with their direct parent in AST instead of all AST-nodes within their scope.
For Variant 2, the AST code sequence tokens are fully connected. Isolation among tokens is lost, which leads to the loss of structural information. The results show the effectiveness of isolation and structural information introduced by SCG. We discuss possible future directions for improvement in Section \ref{sec:dis_graph}.

Our experiments can provide an explanation as to why \citet{NCS} did not improve performance when using SBT \citep{DeepCom}. In an SBT the AST-nodes can be regarded as fully connected. The structural information from the AST graph is not preserved, which is similar to Variant 2. Moreover, SBT introduces redundant interactions between AST-nodes and token-nodes similar to Variant 1. However, Variant 2 still outperforms a vanilla Transformer by 0.43 on BLEU, 0.47 on METEOR, and 0.78 on ROUGE-L. 


\subsection{Expert User Study}
\label{sec:exp_survey}

We make a survey publicly available through an email list to university students and faculty in the Computer Science department as well as virtual anonymous communities on topics related to programming. We ask each participant to select the description that best match the functionality of a code snippet. The survey question is meant to evaluate two modalities of human perceived quality (comprehension of summary) as well as technical accuracy of the summary. For each snippet we provide a user with the code summary generated by GN-Transformer, \citet{GREAT}, Transformer and Transformer (full) by \citet{NCS}. We use an online survey tool\footnote{\url{https://www.qualtrics.com/}} to track and verify the responses. We ask participants for demographic information followed by a randomized selection of code snippets and collect 790 responses from 330 participants. 
Example of a survey question can be found in Appendix Figure A1.

Figure \ref{fig:survey} shows a bar chart with the frequency at which each model was selected. Test of proportions show statistical significant difference between the three models with $\textit{p-value}<0.05$. The survey results agree with the reported automatic metrics (BLEU, METEOR,ROUGE-L) and correlate with the human perceived quality and accuracy. GN-Transformer outperform all other baselines and an annotator score of 37\% compared to 17\% 20\% and 26\% for ``GREAT'', ``Transformer'' and ``Transformer (full)'' respectively. Randomly uniform demographic distribution among participants for every model show that there was no influence by the level of experience, expertise or educational background. 

We conclude that an incremental improvement in automatic metrics can have a significant difference in human perceived quality and accuracy.

\begin{table}[!htb]
\caption{Experimental results on positional encoding. APE/RPE (token) means only applied APE/RPE on token-nodes. Parameter number not including the embedding layers.}
\begin{tabular}{cccc}
\hline
\textbf{Models} & \textbf{BLEU}  & \textbf{METEOR} & \textbf{ROUGE-L} \\ \hline
\multicolumn{4}{c}{\textbf{Java}}                                     \\ \hline
base            & \textbf{45.48} & 26.91           & 55.29            \\
APE             & 44.18          & 25.68           & 54.41            \\
APE (token)     & 44.42          & 25.81           & 54.59            \\
RPE            & 45.32          & \textbf{27.08}  & \textbf{55.33}   \\
RPE (token)    & 45.07          & 26.58           & 55.12            \\ \hline
\multicolumn{4}{c}{\textbf{Python}}                                   \\ \hline
base            & \textbf{33.33} & \textbf{19.67}  & 46.57            \\
APE             & 29.40          & 17.08           & 43.17            \\
APE (token)     & 31.40          & 18.37           & 45.10            \\
RPE            & 32.49          & 19.62           & 46.30            \\
RPE (token)    & 31.55          & 19.09           & 45.70            \\ \hline
\end{tabular}
\label{table:rpe_results}
\end{table}

\section{Discussion}
\subsection{Discussion on graph structure}
\label{sec:dis_graph}


\begin{figure}[h]
\centering
\includegraphics[width=0.75\linewidth]{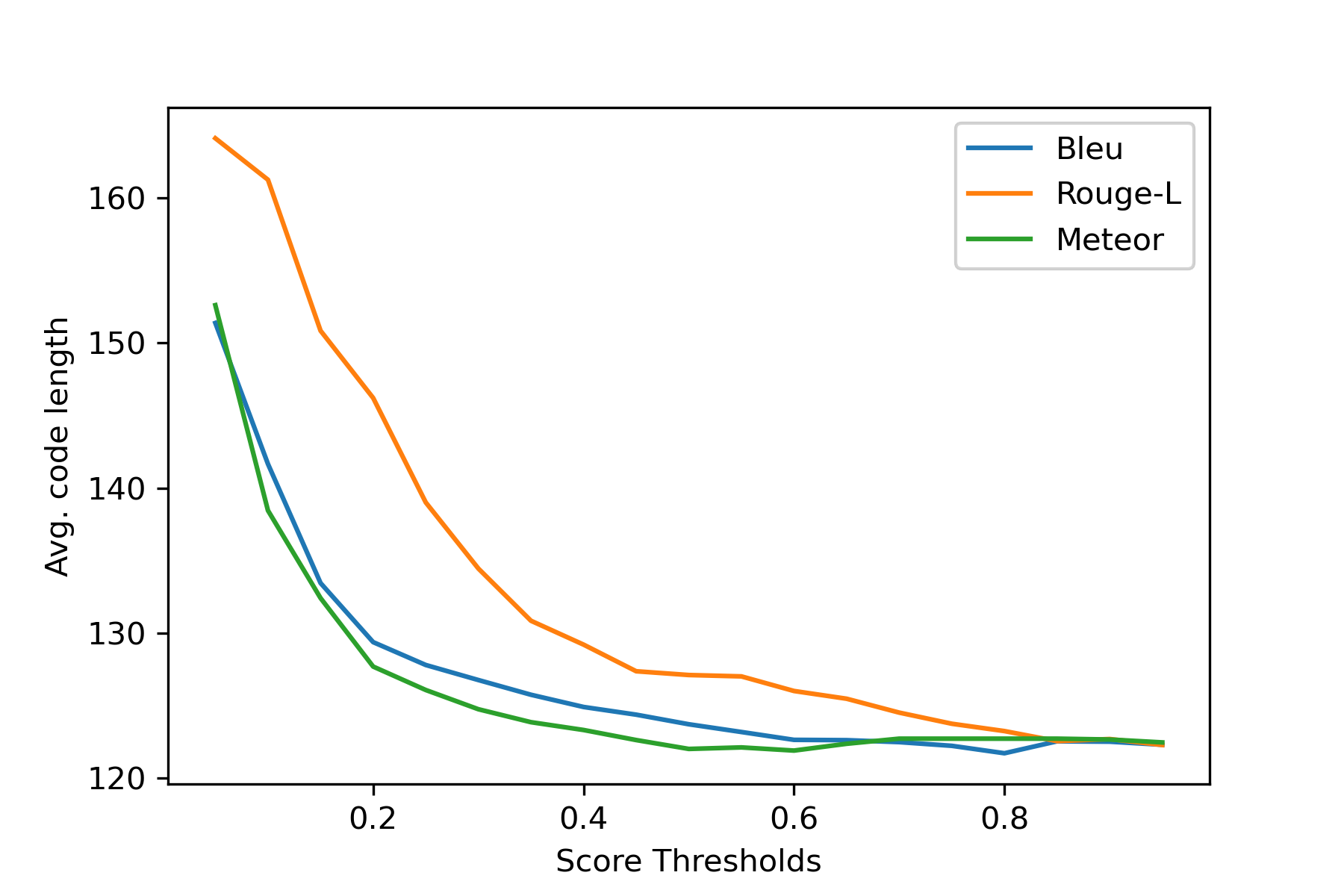}
\centering
\caption{The average code length of the test set of Java dataset on thresholds under different sentence BLEU, ROUGE-L, and METEOR score. The performance of the model degrades across all metrics when larger code snippets are used. } 
\label{figure:threshold}
\end{figure}

An SCG is built on top of an AST structure generated by a parser. Thus an SCG does not introduce any additional assumptions and uses only the grammatical knowledge from the compiler. 
While SCG is a noise-free representation, the inductive bias is introduced by the static edge relationships of an AST. This inflexibility can pose an obstacle in learning more complex relationships from code. 

Qualitatively, we observed two factors of SCG that affect the performance in the code summarization task. The first factor is the long-range dependencies present in code. It is difficult to pass information between nodes separated by long paths in SCG.
As discussed in \ref{sec:gn}, each GN block only implements one round of information propagation.
In Figure \ref{figure:ast_scg}(b), it would require three rounds for information to propagate from the leaf node `a' to root node `AssignExpr'. This is equivalent to at least three GN-Transformer blocks used sequentially. Thus, it is difficult for information to propagate between nodes in a large AST which may require a large number of GN-Transformer blocks. The long-term dependency problem could be ameliorated by designing additional direct edges between AST-nodes and token-nodes. 

The second factor that affects performance is the isolation between token-nodes. Our ablation study of Variant 2 shows the advantage of introducing isolation between the token-nodes when compared to fully connected nodes. Neither can be the optimal solution. In a fully connected source code sequence, a token-node will have as neighbors the preceding and proceeding tokens. In an SCG token-nodes can interact only indirectly with each other through AST-nodes. Consider the expression `5*c' in Figure \ref{figure:ast_scg}(b) as an example. There is no direct edge between token-nodes `5' and `c'. For information to propagate between the two nodes, it has to interact with AST-node `AdditiveExpr'. 

Longer code contains long-range dependencies and further isolated token-nodes that correspond to our two observations on the SCG. We analyzed the BLEU, ROUGE-L, and METEOR scores achieved by the model in relation to the code snippet length on a test set of the Java dataset, the results can be found in Figure \ref{figure:threshold}. The results show that introducing interaction between nodes without discarding isolation is an important future direction for our work.

\subsection{Positional encoding for graph}
\label{sec:pos_encoding}
Unlike sequence models like RNN and LSTM, Transformers do not have an intrinsic mechanism to encode the positional information. There are positional encoding methods proposed in the vanilla Transformer. However, the implementation of such methods can not be expanded directly to a graph. The nodes and edges of a graph are permutation invariant. It is not trivial representing position information in the graph. We did two experiments on positional encodings for a graph. The first experiment uses the \textit{Absolute Positional Embedding} directly applied as in \citep{Transformer}. The second experiment uses a \textit{Relative Positional Encoding}  based on PGNN \cite{PGNN}. 

Table \ref{table:rpe_results} shows the experimental results.
We applied a learnable Absolute Positional Embedding (\textbf{APE}) layer, we use the summation of APE and node embeddings fetched from the input embedding layer as the input to the encoder. We tested APE on all nodes and APE (token) only on token-nodes.

For the experiments on Relative Positional Encoding (\textbf{RPE}) on our model, we had to adapt the original definition for sequence models to that of a graph representation. When RPE \citep{RPE} is applied to sequences, it requires the sequence nodes to be fully connected since the relative position is modeled in relation to an attribute on a direct edge between two token nodes, which is not the case for SCG. 
Instead, we apply a two-layered PGNN \citep{PGNN} that learns relative positional information on a graph to implement RPE for each node. We used an APE as the input to PGNN. We set an embedding size of 512 and a fixed number of 6 anchor sets with 2 copies each instead of an adjusted anchor set number in \citet{PGNN}. 
We concatenate RPE and node embeddings fetched from the input embedding layer. 
Experiments noted as RPE (token) apply RPE only on toke-nodes with padding on AST-nodes.

The results show that APE will harm the performance of both the Java and Python datasets. When applying APE only in token-nodes, the degradation is minor. We hypothesize that APE is not useful for AST-nodes. APE in token-nodes represents their absolute positions in the input sequence. On the contrary, the positions of AST-nodes in the input sequence should be the scope in relation to the token-node. The absolute positional information is not useful in a standard SCG for the AST-nodes. 

The results for RPE are also not promising in the interpretation we made for graph representations. We chose a fixed number of anchor sets for all graphs. In \citet{PGNN}, they dynamically chose anchor set numbers for different sizes of graphs. We hypothesize that this approach is limited by the insufficient anchor set numbers in large graphs. Consider that the average node number of the Python dataset is 70.10 compared to 50.30 in the Java dataset. This can explain the reason RPE improves the performance in the Java dataset but there is performance degradation for the Python dataset when compared with a base model. 

The results motivate future work to find a suitable positional representation in the context of Graph Networks.


\section{Conclusion}
\label{others}

In this paper, we analyze the fusion of a sequence and a graph from a novel perspective of Graph Networks. We propose a GN-Transformer and Syntax-Code Graph. Our method achieve state-of-the-art in two code summarization datasets. We perform experiments on hyperparameters, model structure, and graph structure. We perform the largest to date expert-user study on human perceived quality and accuracy for a code summarization dataset to corroborate with automatic metrics. 
Future work include finding a method to ameliorate the long-term dependency problem of SCG structure without introducing noise and an effective way of representing the positional information of nodes in a graph. Due to the similarity with Transformer, ideas like masked pretraining can also be interpreted and implemented in the context of Graph Networks. A decoder interpretation for graphs could also improve performance. In our method, we discard AST-node embeddings thus losing some structural information. Our method could be used in any domain where there is a duality of a sequence and graph representations.

\bibliographystyle{ACM-Reference-Format}
\bibliography{main}

\clearpage

~\\
\begin{lstlisting} [language = Java,
        keywordstyle=\color{blue!70},
        commentstyle=\color{red!50!green!50!blue!50},
        frame=topline|bottomline|leftline|rightline,
        rulesepcolor=\color{red!20!green!20!blue!20},
        basicstyle=\small ]
protected static void quickSort(Instances insts,int[] indices,int attidx,int left,int right){
    if (left < right) {
        int middle=partition(insts,indices,attidx,left,right);
        quickSort(insts,indices,attidx,left,middle);
        quickSort(insts,indices,attidx,middle + 1,right);
    }
}
\end{lstlisting} 

\textbf{Reference:} sorts the instances according to the given attribute/dimension. the sorting is done on the master index array and not on the actual instances object.

\textbf{Ours:} sorts the specified range of the array using the specified items

\textbf{Transformer:} src the ordinal field array into ascending order

\textbf{Transformer (full):} sorts the specified range of the array using the given workspace array .


~\\
\begin{lstlisting}[language = Python,
        keywordstyle=\color{blue!70},
        commentstyle=\color{red!50!green!50!blue!50},
        frame=topline|bottomline|leftline|rightline,
        rulesepcolor=\color{red!20!green!20!blue!20},
        basicstyle=\small]
def is_power2(num): 
    return (isinstance(num, numbers.Integral) and (num > 0) and (not (num & (num - 1))))
\end{lstlisting}

\textbf{Reference:} test if num is a positive integer power of 2 .

\textbf{Ours:} return true if the power of 2 .

\textbf{Transformer:} returns true if and number is a user-defined power .

\textbf{Transformer (full):} return whether or not the argument is a power .

\newpage
~\\
\begin{lstlisting}[language = Python,
        keywordstyle=\color{blue!70},
        commentstyle=\color{red!50!green!50!blue!50},
        frame=topline|bottomline|leftline|rightline,
        rulesepcolor=\color{red!20!green!20!blue!20},
        basicstyle=\small]
def _mergeOptions(inputOptions, overrideOptions): 
    if inputOptions.pickledOptions: 
        try: 
            inputOptions = base64unpickle(inputOptions.pickledOptions) 
        except Exception as ex: 
            errMsg = (``provided invalid value `%s' for option `--pickled-options'" % inputOptions.pickledOptions) 
            errMsg += ((``(`%s')" % ex) if ex.message else `') 
            raise SqlmapSyntaxException(errMsg) 
        if inputOptions.configFile: 
            configFileParser(inputOptions.configFile) 
        if hasattr(inputOptions, `items'): 
            inputOptionsItems = inputOptions.items() 
        else: 
            inputOptionsItems = inputOptions.__dict__.items() 
        for (key, value) in inputOptionsItems: 
            if ((key not in conf) or (value not in (None, False)) or overrideOptions): 
                conf[key] = value 
        for (key, value) in conf.items(): 
            if (value is not None): 
                kb.explicitSettings.add(key) 
        for (key, value) in defaults.items(): 
            if (hasattr(conf, key) and (conf[key] is None)): 
                conf[key] = value 
        _ = {} 
        for (key, value) in os.environ.items(): 
            if key.upper().startswith(
                SQLMAP_ENVIRONMENT_PREFIX): 
                _[key[len(SQLMAP_ENVIRONMENT_PREFIX):]
                    .upper()] = value 
        types_ = {} 
        for group in optDict.keys(): 
            types_.update(optDict[group]) 
        for key in conf: 
            if ((key.upper() in _) and (key in types_)): 
                value = _[key.upper()] 
                if (types_[key] == OPTION_TYPE.BOOLEAN): 
                    try: 
                        value = bool(value) 
                    except ValueError: 
                        value = False 
                elif (types_[key] == OPTION_TYPE.INTEGER): 
                    try: 
                        value = int(value) 
                    except ValueError: 
                        value = 0 
                elif (types_[key] == OPTION_TYPE.FLOAT): 
                    try: 
                        value = float(value) 
                    except ValueError: 
                        value = 0.0 
                conf[key] = value 
        mergedOptions.update(conf)
\end{lstlisting}

\textbf{Reference:} merge command line options with configuration file and default options .

\textbf{Ours:} merges options from a config file .

\textbf{Transformer:} merges all of the data used into an option .

\textbf{Transformer (full):} loads configuration attributes and add attributes .

~\\
\begin{lstlisting}[language = Python,
        keywordstyle=\color{blue!70},
        commentstyle=\color{red!50!green!50!blue!50},
        frame=topline|bottomline|leftline|rightline,
        rulesepcolor=\color{red!20!green!20!blue!20},
        basicstyle=\small]
def LoadFromString(yaml_doc, product_yaml_key, required_client_values, optional_product_values): 
    if (_PY_VERSION_MAJOR == 2): 
        if ((_PY_VERSION_MINOR == 7) and (_PY_VERSION_MICRO < 9)): 
            _logger.warning(_DEPRECATED_VERSION_TEMPLATE, _PY_VERSION_MAJOR, _PY_VERSION_MINOR, _PY_VERSION_MICRO) 
        elif (_PY_VERSION_MINOR < 7): 
            _logger.warning(_DEPRECATED_VERSION_TEMPLATE, _PY_VERSION_MAJOR, _PY_VERSION_MINOR, _PY_VERSION_MICRO) 
    data = (yaml.safe_load(yaml_doc) or {}) 
    try: 
        product_data = data[product_yaml_key] 
    except KeyError: 
        raise googleads.errors.GoogleAdsValueError((`The ``%s" configuration is missing' % (product_yaml_key,))) 
    if (not isinstance(product_data, dict)): 
        raise googleads.errors.GoogleAdsValueError((`The ``%s" configuration is empty or invalid' % (product_yaml_key,))) 
    IncludeUtilitiesInUserAgent(
        data.get(_UTILITY_REGISTER_YAML_KEY, True))
    original_keys = list(product_data.keys()) 
    client_kwargs = {} 
    try: 
        for key in required_client_values: 
            client_kwargs[key] = product_data[key] 
            del product_data[key] 
    except KeyError: 
        raise googleads.errors.GoogleAdsValueError((`Some of the required values are missing. Required values are: %s, actual values are %s' % (required_client_values, original_keys))) 
    proxy_config_data = data.get(_PROXY_CONFIG_KEY, {}) 
    proxy_config = _ExtractProxyConfig(product_yaml_key, proxy_config_data) 
    client_kwargs[`proxy_config'] = proxy_config 
    client_kwargs[`oauth2_client'] = _ExtractOAuth2Client(product_yaml_key, product_data, proxy_config) 
    client_kwargs[ENABLE_COMPRESSION_KEY] = data.get(ENABLE_COMPRESSION_KEY, False) 
    for value in optional_product_values: 
        if (value in product_data): 
            client_kwargs[value] = product_data[value] 
            del product_data[value] 
    if product_data: 
        warnings.warn((`Could not recognize the following keys: %s. They were ignored.' % (product_data,)), stacklevel=3) 
    return client_kwargs
\end{lstlisting}

\textbf{Reference:} loads the data necessary for instantiating a client from file storage .

\textbf{Ours:} loads the header data necessary for instantiating .

\textbf{Transformer:} loads a data necessary for credit two types .

\textbf{Transformer (full):} loads key: data from a loader\_context .

\newpage
\renewcommand{\thefigure}{A\arabic{figure}}

\setcounter{figure}{0}

\begin{figure*}[]
\centering
\includegraphics[width=1\linewidth]{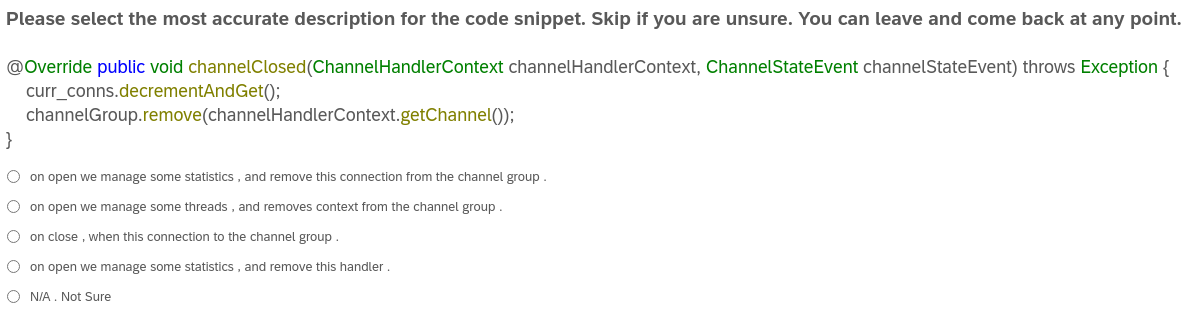}
\centering
\caption{Randomly selected expert-user question sample.} 
\label{app:survey}
\end{figure*}

\clearpage
\section{Attention Visualization for Syntax-Code Graph}

The attention visualization for this short program is shown here, we could get more insights about how each nodes paid attention to their neighboring nodes:

\begin{lstlisting} [language = Java,
        keywordstyle=\color{blue!70},
        commentstyle=\color{red!50!green!50!blue!50},
        frame=topline|bottomline|leftline|rightline,
        rulesepcolor=\color{red!20!green!20!blue!20},
        basicstyle=\small ]
public StartListener(Object resource) {
    resource=resource;
}
\end{lstlisting}

The figures show attention visualization for standard SCG. We marked the edges within AST-nodes by blue, edges between AST-nodes and token-nodes by red. Each figure shows the attention visualization of the nodes in each layer of AST. 

\newpage
\begin{figure*}[]
\begin{center}
\includegraphics[]{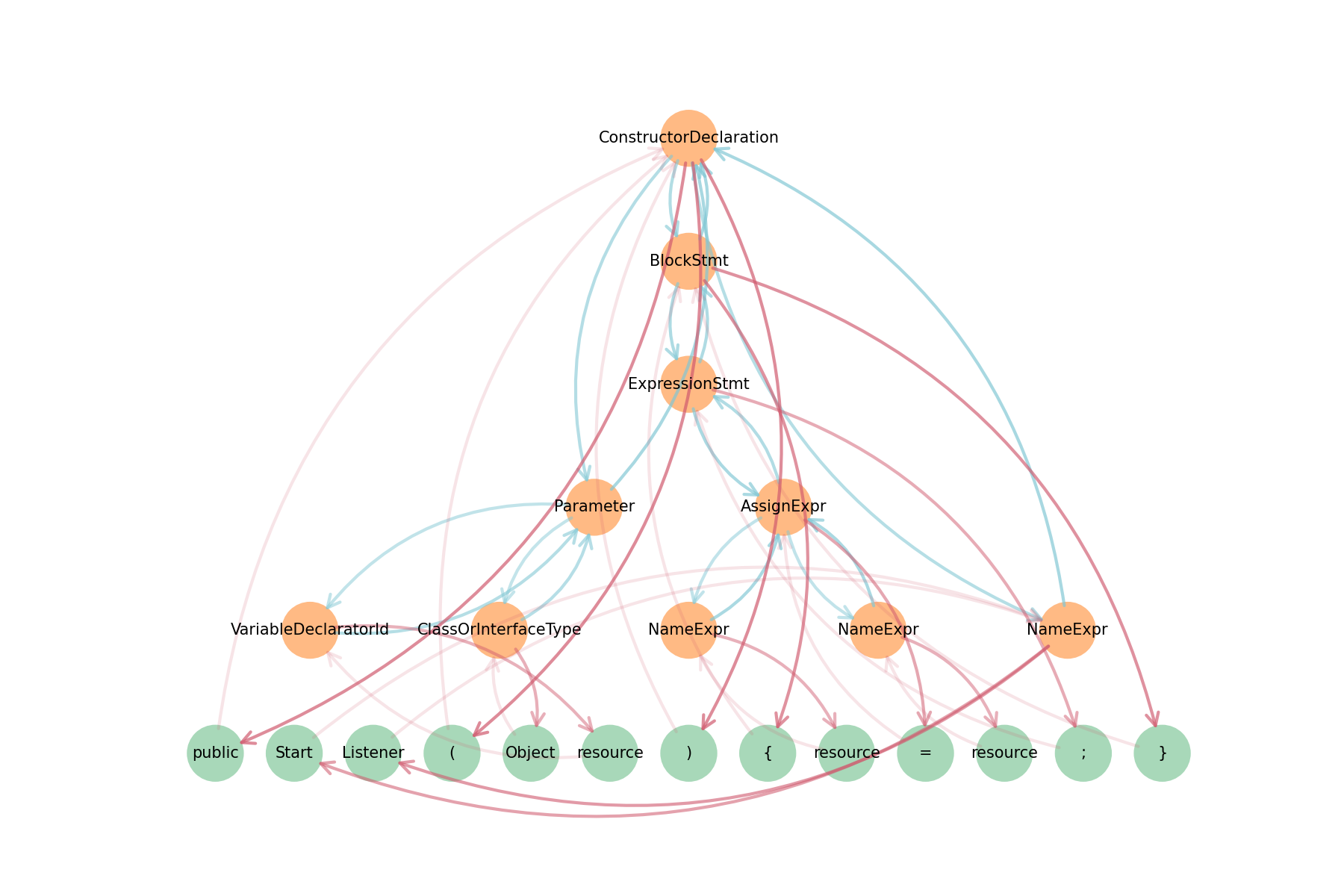} 
\end{center}
\end{figure*}

\begin{figure*}
\begin{center}
\includegraphics[]{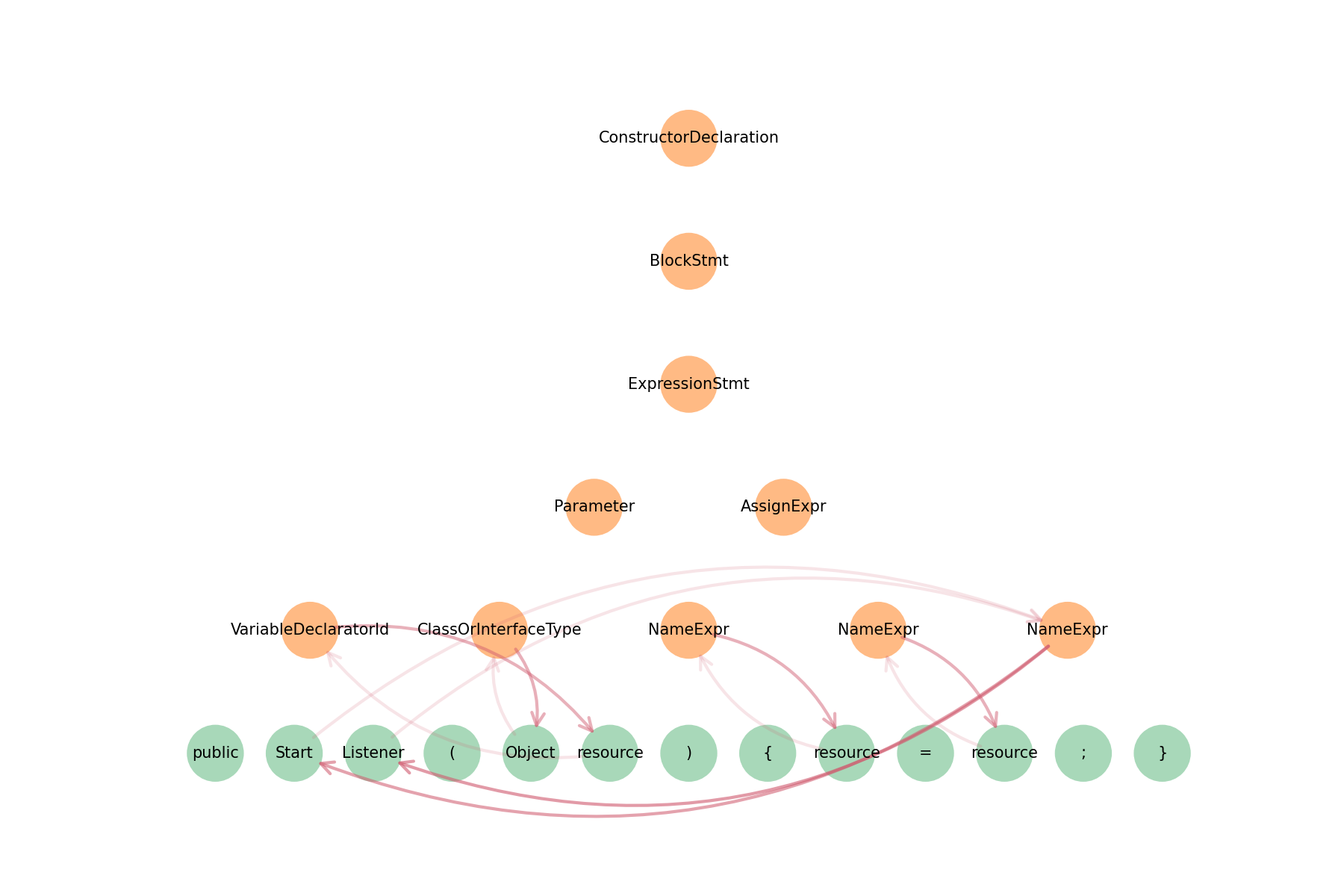} 
\end{center}
\end{figure*}

\begin{figure*}
\begin{center}
\includegraphics[]{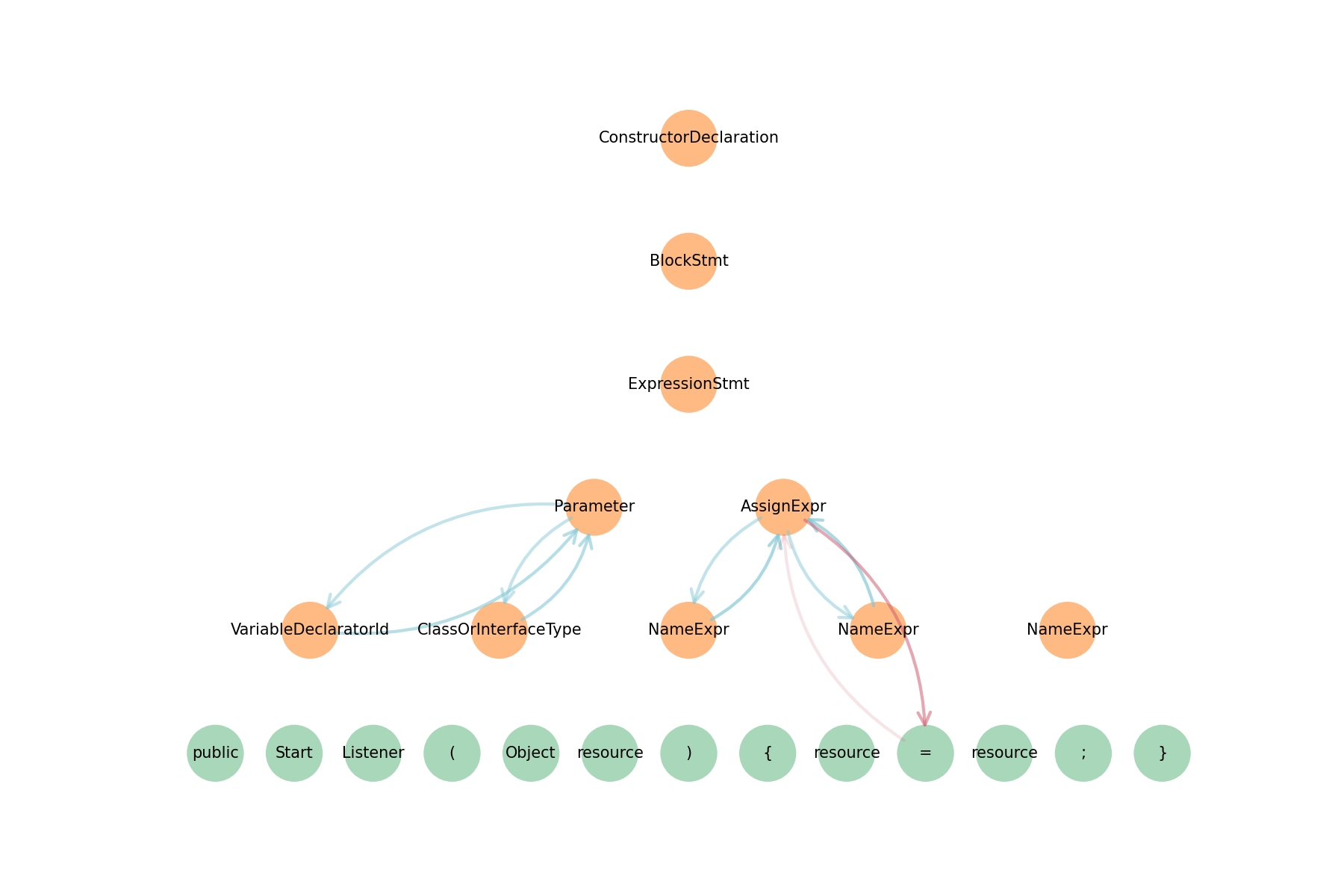} 
\end{center}
\end{figure*}

\begin{figure*}
\begin{center}
\includegraphics[]{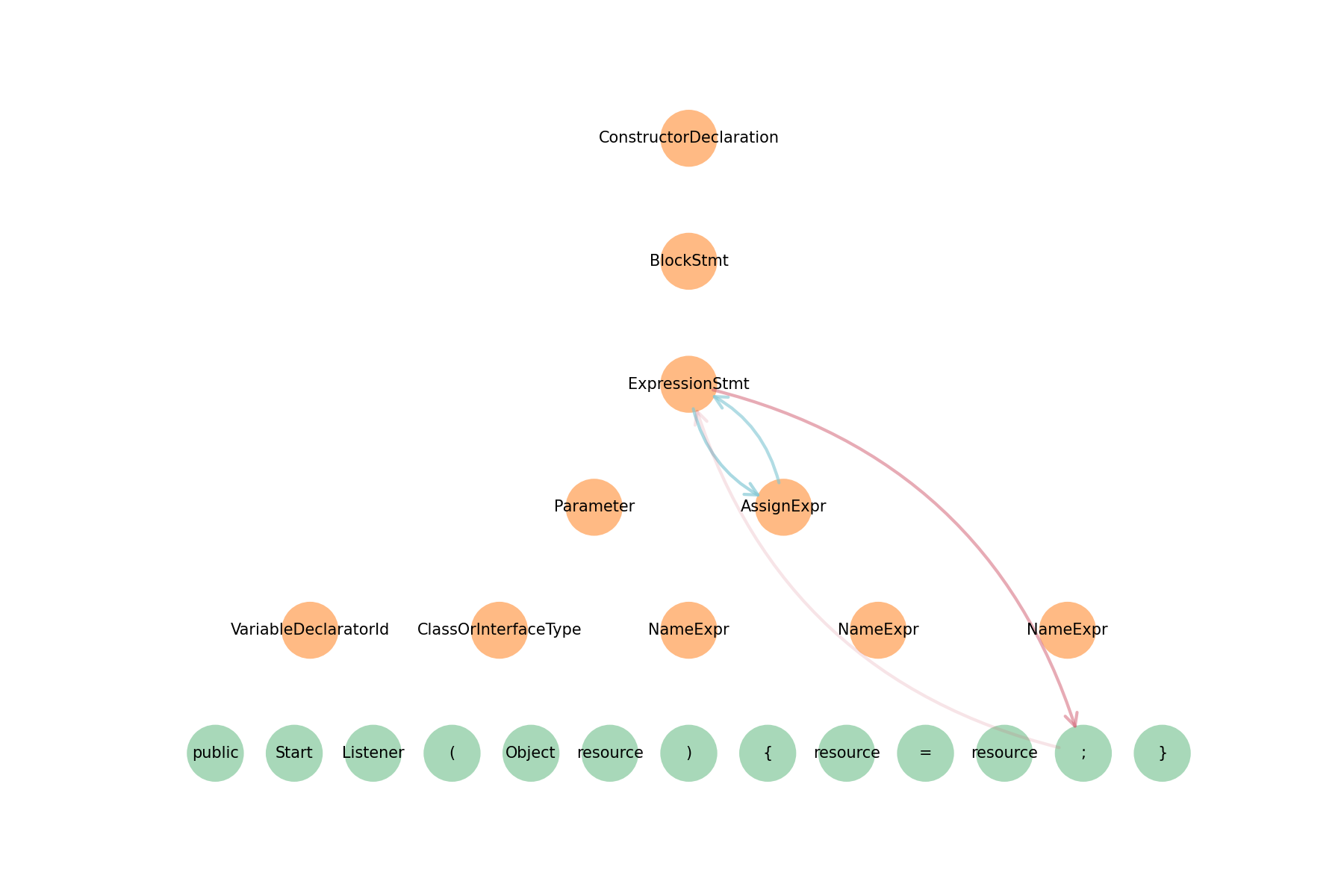} 
\end{center}
\end{figure*}

\begin{figure*}
\begin{center}
\includegraphics[]{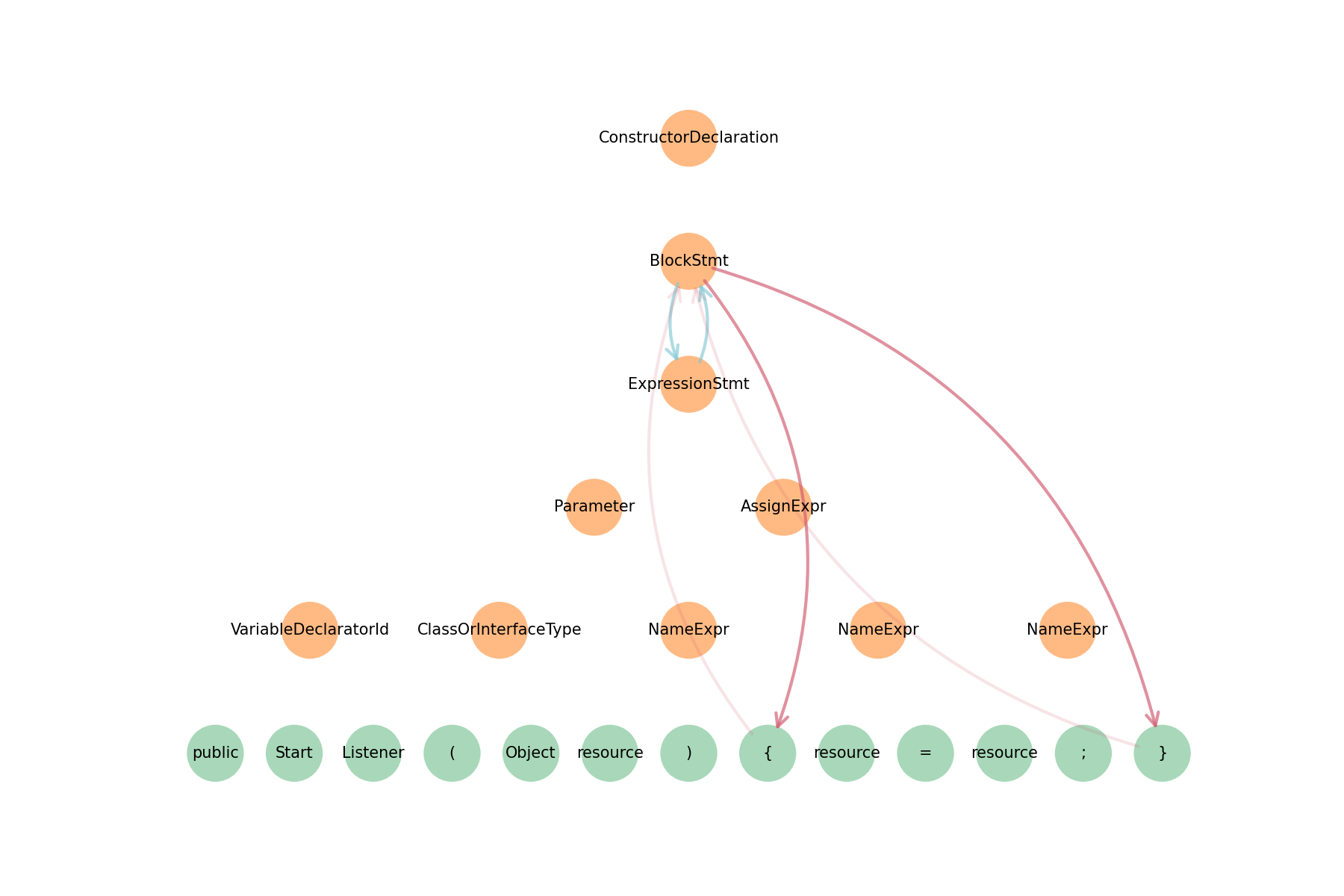} 
\end{center}
\end{figure*}

\begin{figure*}
\begin{center}
\includegraphics[]{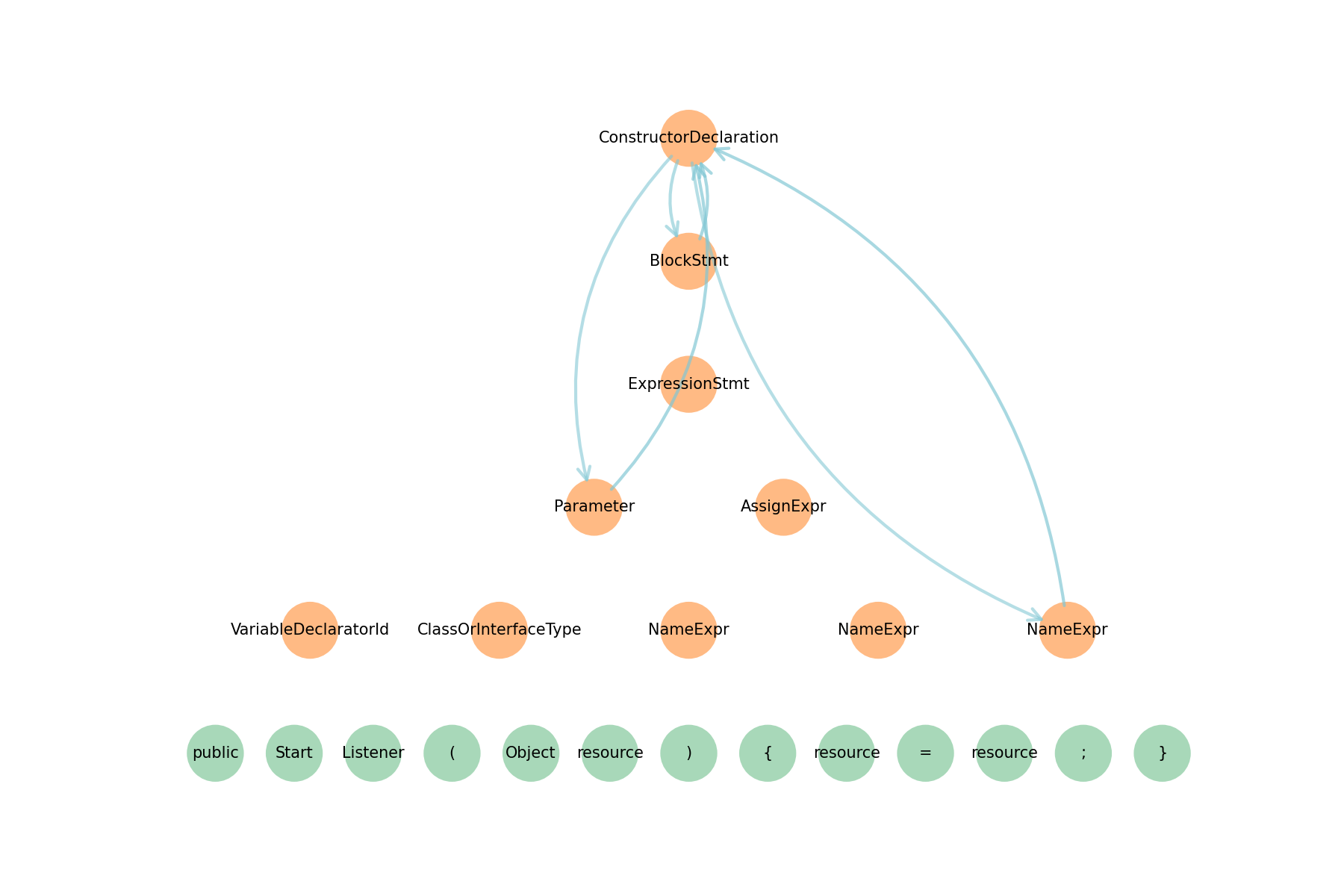} 
\end{center}
\end{figure*}

\end{document}